\documentclass{article}

\usepackage[preprint]{neurips_2026}

\usepackage[utf8]{inputenc}
\usepackage[T1]{fontenc}
\usepackage{hyperref}
\usepackage{url}
\usepackage{booktabs}
\usepackage{amsfonts}
\usepackage{amsmath}
\usepackage{nicefrac}
\usepackage{microtype}
\usepackage{xcolor}
\usepackage{graphicx}
\usepackage{wrapfig}
\usepackage{subcaption}
\usepackage{enumitem}
\usepackage[nameinlink,noabbrev]{cleveref}

\newif\ifshowcoloredits
\showcoloreditsfalse

\definecolor{bccolor}{RGB}{143,0,255}
\definecolor{spcolor}{RGB}{25,95,170}
\definecolor{mgcolor}{RGB}{180,30,140}
\definecolor{jlcolor}{RGB}{200,0,0}
\definecolor{todocolor}{RGB}{200,0,0}
\definecolor{pmiloscolor}{RGB}{170,80,0}
\definecolor{azcolor}{RGB}{220,70,160}

\newcommand{\editcomment}[3]{%
  \ifshowcoloredits
    {\textcolor{#2}{\small [#1: #3]}}%
  \fi
}
\newcommand{\bc}[1]{\editcomment{bc}{bccolor}{#1}}

\newcommand{\mg}[1]{\editcomment{mg}{mgcolor}{#1}}

\newcommand{\todo}[1]{\editcomment{todo}{todocolor}{#1}}

\newcommand{\alicja}[1]{\editcomment{az}{azcolor}{#1}}

\title{What Drives Interactive Improvement from Feedback?}

\newcommand{\affmark}[1]{\textsuperscript{#1}}

\author{%
  \textbf{Bartłomiej Cupiał}\affmark{1,2,*} \qquad
  \textbf{Jan Łojek}\affmark{1} \qquad
  \textbf{Mikołaj Garstecki}\affmark{1} \\
  \textbf{Szymon Pobłocki}\affmark{1} \qquad
  \textbf{Alicja Ziarko}\affmark{1,2,3} \qquad
  \textbf{Piotr Miłoś}\affmark{1,4,5}
}

\begin{document}

\newcommand{\acc}{\mathrm{acc}}
\newcommand{\gain}{\mathrm{gain}}
\newcommand{\ngain}{\mathrm{ngain}}
\newcommand{\auc}{\mathrm{auc}}

\maketitle

\begingroup
\makeatletter
\renewcommand\@makefntext[1]{\noindent#1}
\makeatother

\footnotetext{Accepted to ``RLxF: Reinforcement Learning from World Feedback''.}
\footnotetext{
    \affmark{*}Corresponding author: \texttt{bartlomiej.cupial@gmail.com}. 
    \affmark{1}University of Warsaw \quad
    \affmark{2}AKCES NCBR \quad
    \affmark{3}Princeton University \quad
    \affmark{4}Mistral AI \quad
    \affmark{5}Institute of Mathematics, Polish Academy of Sciences \quad
}
\endgroup

\begin{abstract}
We study when natural-language feedback produces improvement beyond the gains obtainable from repeated attempts alone. In multi-turn language agent setting, higher final accuracy can reflect useful feedback, but it can also arise from resampling, format correction, or additional test-time computation. To separate these effects, we introduce a controlled student-teacher protocol across Omni-MATH, Codeforces, BBEH Linguini, and ARC-AGI1, evaluating thirteen open-weight models in both student and teacher roles. We compare external feedback, self-feedback, and unguided self-refinement, while varying interaction history, task difficulty, and teacher access to privileged task information. 
Across settings, we find that multi-turn improvement is often not evidence of feedback use: self-generated feedback adds little beyond unguided self-refinement, whereas the strongest external teachers produce substantially larger feedback-specific gains, suggesting that useful feedback must provide guidance beyond generic retry.
Dense student-teacher interaction matrices further show that interactive gains are driven more by the student’s ability to use feedback than by the teacher’s identity, although teacher choice remains important for a fixed student. 
These results suggest that feedback-based agents should be evaluated against repeated-attempt baselines, and that ability to act on feedback, not merely feedback availability, is a central bottleneck for interactive improvement. 
We release our controlled student-teacher evaluation framework at https://j-lojek.github.io/feedback-generation-is-a-bottleneck/.
\end{abstract}

\section{Introduction}
\label{sec:introduction}

\begin{wrapfigure}[17]{r}{0.52\linewidth}
\vspace{-4em}
\centering
\includegraphics[width=\linewidth]{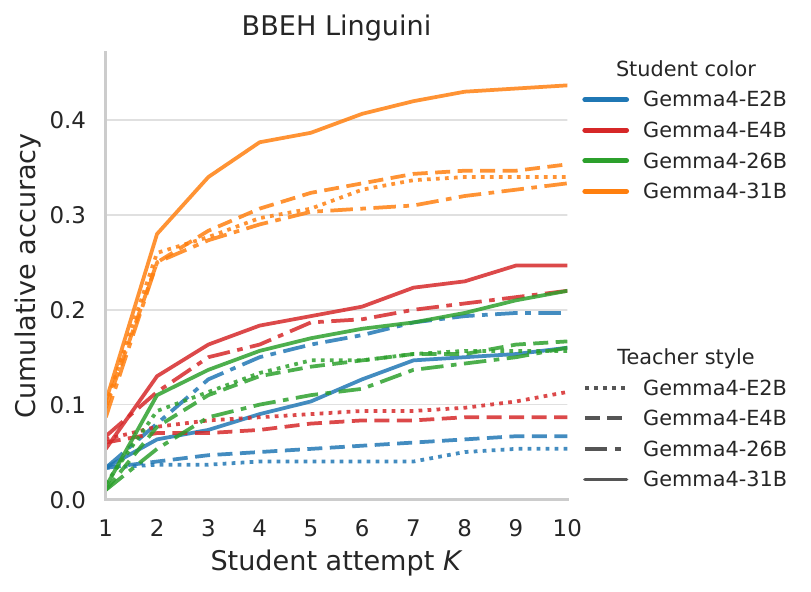}
\caption{
\textbf{Multi-turn trajectories separate mostly by student.}
While stronger teachers like Gemma4-31B provide noticeable boosts for highly capable students, the dominant vertical separation between the curves is driven by student rather than teacher.} 
\label{fig:teaser-gemma4-bbeh-linguini}
\vspace{-11em}
\end{wrapfigure}

Language models (LMs) are increasingly deployed as agents that act, observe, 
and revise rather than as systems that produce a single isolated answer. 
Coding agents react to tests and error traces, 
computer-use agents observe changing interface states, and assistants
revise plans in response to user corrections or clarifications. Across these
settings, feedback can do more than mark success or failure: it can explain what
went wrong, identify missing constraints, and suggest a path to recovery.
Improving from feedback is therefore becoming an important capability in its own
right: an agent must recognize which parts of the feedback are actionable,
preserve the parts of its attempt that are already correct, and revise the parts
that caused failure.

Studying improvement from feedback requires separating several mechanisms that are usually entangled. A model may improve over turns because it receives useful corrective information, but it may also improve simply because it samples another attempt with more computation. When feedback does help, the limiting factor may lie on either side of the interaction: the teacher may fail to identify the relevant error, or the student may fail to translate a correct diagnosis into a better solution. We therefore distinguish feedback generation from improving from feedback by evaluating separate student and teacher models under controlled interaction histories, teacher-information conditions, and repeated-attempt baselines. This lets us ask not only whether feedback improves performance, but when it improves performance beyond self-refinement, and which part of the feedback system constrains that improvement.


We answer these questions in a student-teacher protocol on verifiable tasks. 
In each episode, a student model attempts a problem. If the attempt 
is incorrect, a teacher model provides natural-language feedback, and the
student revises its answer. The verifier makes every attempt measurable,
while the student-teacher split lets us vary the student, the teacher, 
the interaction budget, the visible history, and the teacher's access
to privileged task information. We instantiate this evaluation across four
verifiable reasoning environments: Omni-MATH, Codeforces, BBEH Linguini, and
ARC-AGI1, and run student-teacher interaction grids in which each model acts in
both roles.

Across these settings, we find that multi-turn improvement is often not evidence of feedback use: repeated attempts explain a large share of the gain, self-feedback improves over generic retry inconsistently, the strongest feedback-specific improvements come from high-quality external teachers, and the student's ability to use feedback explains more variation than the teacher's identity.

\bc{ figure out where to put this:
"We find that feedback gains are limited unless feedback comes from substantially stronger teachers. Self-generated feedback often adds little beyond unguided self-refinement, suggesting that producing useful feedback is itself a difficult capability rather than an automatic consequence of attempting the task. This contrasts with the simple generator-verifier intuition: to improve a student, a teacher must not only recognize that an answer is wrong, but diagnose the student’s specific error and express a correction the student can act on."}

In summary, our contributions are as follows:
\begin{itemize}
    \item We introduce a controlled zero-shot student-teacher evaluation for natural-language feedback across four verifiable reasoning environments.
    \item We show that multi-turn improvement is not necessarily feedback use: self-feedback adds little beyond self-refinement, while the best external teachers add substantial feedback-specific gains.
    \item We find that interactive performance is primarily student-dependent: student identity explains substantially more gain variation than teacher identity, although teacher choice can matter for a fixed student.
    \item We show that longer histories and privileged teacher information are selective interventions rather than default improvements, with benefits depending on task and model capability.
\end{itemize}

\section{Related Work}
\label{sec:related-work}

\paragraph{Self-refinement and repeated sampling.}

Large language models are increasingly used as interactive agents that solve tasks over multiple turns rather than through a single generation~\citep{wang2024mintevaluatingllmsmultiturn}. In such settings, models must follow sequential instructions and incorporate new feedback across turns~\citep{laban2025llmslostmultiturnconversation}.
However, multi-turn improvement does not necessarily imply that the model has used feedback: later attempts also receive more test-time computation and additional samples.
This distinction is especially important because additional computation is known to improve performance through mechanisms such as chain-of-thought prompting~\citep{wei2023chainofthoughtpromptingelicitsreasoning}, and Best-of-$N$ sampling~\citep{wang2023selfconsistencyimproveschainthought}.
Related agent systems such as Reflexion use verbal feedback or memories of past failures to improve future attempts~\citep{shinn2023reflexionlanguageagentsverbal}. 
Our work builds on this line of work by explicitly comparing feedback-conditioned improvement against unguided self-refinement, allowing us to estimate how much of the observed gain is feedback-specific rather than a consequence of repeated attempts.

\paragraph{Post-training LMs with teacher feedback.}
Recent post-training methods increasingly frame model improvement as a student-teacher problem, where a student model learns from signals produced by a stronger teacher, verifier, or environment. Standard approaches include supervised fine-tuning, preference optimization, reinforcement learning from AI feedback, and reinforcement learning with verifiable rewards~\citep{ouyang2022traininglanguagemodelsfollow,lee2024rlaifvsrlhfscaling,rafailov2024directpreferenceoptimizationlanguage,Guo_2025,shao2024deepseekmathpushinglimitsmathematical}. These methods are effective, but their feedback signals are often demonstrations, pairwise preferences, scalar rewards, or outcome-level verification. Such signals can improve behavior, but they provide limited information about which part of a student's attempted solution 
\begin{figure}[h]
\centering
\includegraphics[width=0.85\linewidth]{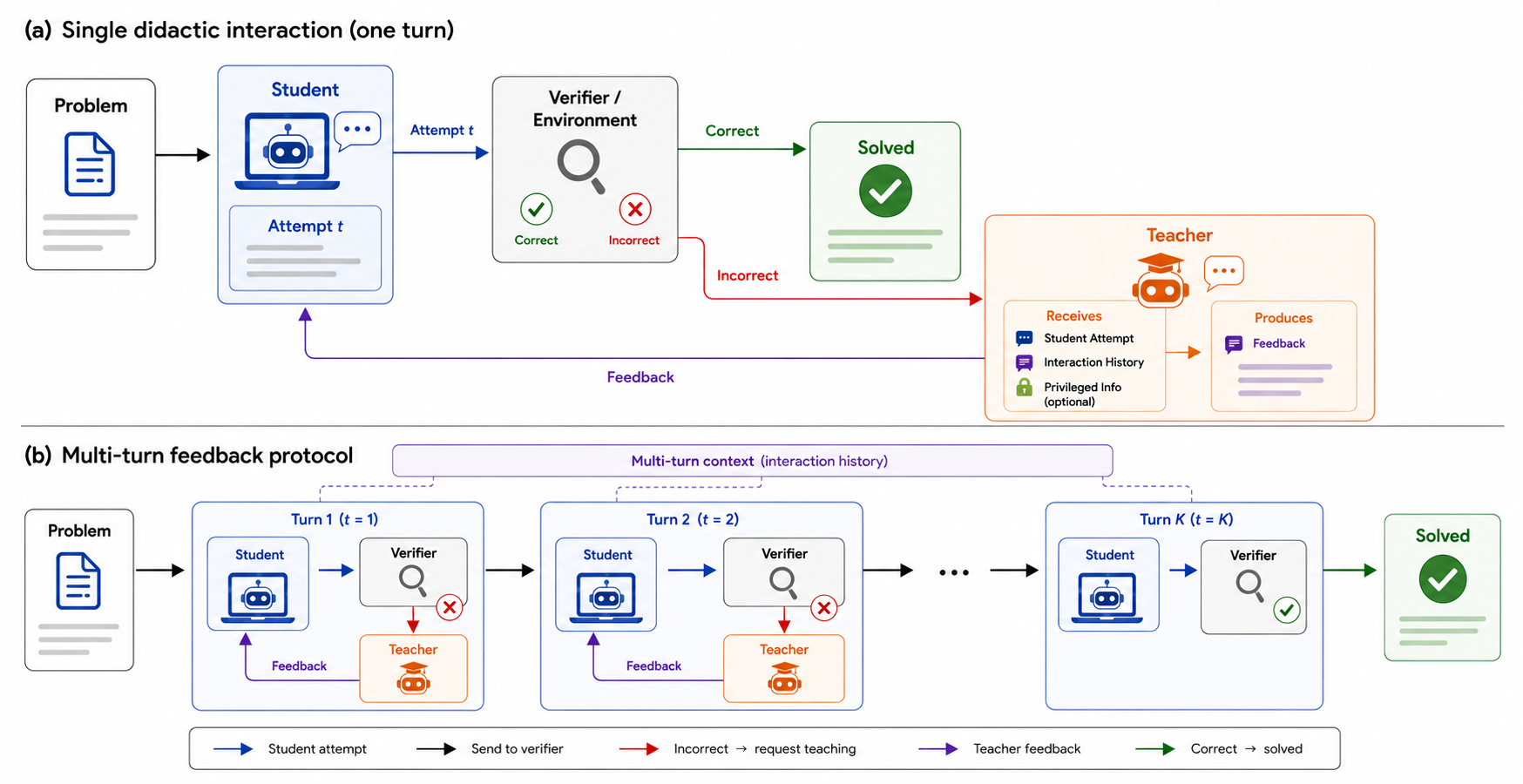}
\caption{\textbf{(Top) Single Didactic Interaction.} We transform single-turn
problems into multi-turn didactic interactions. A teacher model, conditioned on optionally privileged information
(e.g., the ground-truth solution), provides natural language feedback to a student model without
revealing the final answer, guiding it to correct its errors. 
\textbf{(Bottom) Multi-turn feedback protocol.} Each failed student attempt is converted into teacher feedback for the next attempt. This cycle repeats until the predefined maximum number of turns is reached or the student correctly solves the problem.
}
\label{fig:experimental-setup}

\end{figure}
caused failure. This limitation has motivated recent work on 
denser teacher supervision, including on-policy 
distillation, where trajectories are sampled from the student and a teacher provides supervision on the  student-generated states~\citep{lu2025onpolicydistillation}. It has also motivated work on synthetic and interactive data generation, especially as high-quality human-generated training data becomes increasingly 
limited~\citep{villalobos2024willdatarun,kessler2026activesyntheticdatageneration}. In verifiable domains, natural-language feedback can be used to construct multi-turn training trajectories or improve interactive learning from feedback~\citep{klissarov2026interactivefeedback,cook2026learninglearnlanguagefeedback}. Related work has also used natural-language feedback as a fine-tuning signal or as part of reinforcement-learning-style updates~\citep{choudhury2024betterteacherllmagents,feng2024naturallanguagereinforcementlearning}. Our work studies the test-time analogue of this post-training shift: a teacher observes a student's failed attempt and provides diagnostic feedback, and we ask when this feedback improves performance beyond repeated attempts, whether stronger teachers provide more useful corrections, and whether the main bottleneck lies in feedback generation or student uptake.

\paragraph{Feedback generation and tutoring.}
Several works study the quality of feedback itself. \citep{xu2025provablylearninglanguagefeedback} identify conditions under which language feedback can support no-regret learning, including unbiased feedback, the agent's ability to translate feedback into evaluations of candidate hypotheses, and sufficient representational coverage of the true environment. In mathematical reasoning, \citep{li-etal-2024-evaluating-mathematical} evaluate whether models can detect and correct errors in flawed reasoning traces, while \citep{gupta2025finalanswersevaluatinglarge} analyze the tutoring behavior of GPT-family models beyond final-answer correctness. These works focus on whether feedback is valid, helpful, or pedagogically appropriate. Our work complements them by evaluating the full student-teacher interaction: a teacher may generate useful feedback, but improvement also depends on whether the student can act on it.

\section{Experimental Setup}
\label{sec:experimental-setup}
In this section we describe our experimental setup. 
\paragraph{Environments.}
We restrict our evaluation to environments, where for each problem and solution pair, it is possible to verify whether the given solution is correct.
This assumption is often made in the literature \alicja{TODO cite}.
To make our results robust, we consider four challenging environments, 
that focus on different domains and test for different abilities: \textbf{Omni-MATH} \citep{gao2024omnimath} Olympiad-level mathematical reasoning benchmark. \textbf{Codeforces} \citep{competitive_coding_benchmark}, a competitive programming benchmark, the \textbf{Linguini} task from \textbf{BIG-Bench Extra Hard (BBEH)} \citep{kazemi2025bbeh}, which specifically challenges the model's ability to learn new rules and \textbf{ARC-AGI} \citep{chollet2019arc}, which presents grid transformation puzzles solvable only by inferring abstract rules from a few examples. \mg{maybe add a visualization of those datasets as figure.}\bc{would add that in the appendix A}

\paragraph{Interaction protocol.}
For investigating the influence of feedback, we consider a realistic scenario of a back-and-forth conversation between a model 
and an entity capable of giving it feedback (for simplicity, we call them student and teacher).
Each episode is a trajectory $\tau=(\text{problem formulation}, s_1, a_1, t_1, \dots, s_n, a_n)$,
where the initial model input contains the problem formulation.
In turn $i$, $s_i$ is the student's output, $a_i$ is the student's answer.
In the case where the task-specific verifier recognized $a_i$ as the wrong answer, $t_i$ is the teacher's feedback on the student's attempt.
Then, the next turn $i+1$ starts, where the student model attempts to solve the problem again, with the added information of the teacher's feedback.
The conversation continues until a maximal $n$ is reached or the student's answer is correct. 
Our interaction protocol is visualized in ~\Cref{fig:experimental-setup}.

\paragraph{Student's input.}
Lead by the intuition that the recent feedback is more important than old feedback, we restrict the student's
input to only $h$ historical turns, where $h$ is a hyperparameter. 
Therefore, the student's input is $(\text{problem formulation}, s_k, a_k, t_k, \dots, s_n, a_n)$,
where $k=\text{max}(n-h, 0)$.

\paragraph{Teacher's input.}
In previous works \citep{klissarov2026interactivefeedback, cook2026learninglearnlanguagefeedback}, the teacher model was granted access to specific task knowledge, denoted as $k_t$. Due to the varying nature of the evaluation environments we have chosen, this additional information in this work have taken different forms depending on dataset used: an in-depth step-by-step solution, a plain final answer, or an error stack trace derived from the student's generated code. Because smaller models are prone to inadvertently leaking the direct solution when given such context, we establish a strict baseline where no additional information is provided to the teacher ($k_t = \text{none}$). However, to better understand the impact of privileged context, we also include ablations that selectively provide this knowledge based on the specific environment. The configurations are as follows: for the Math environment, the teacher receives either no information, the final answer, or the full solution; for the BBEH environment, the knowledge is restricted to either no information or the final answer; and for the Code environment, the teacher receives either no additional information or the complete solution.


\begin{center}
\centering
\includegraphics[width=\linewidth,trim=0 390bp 0 0,clip]{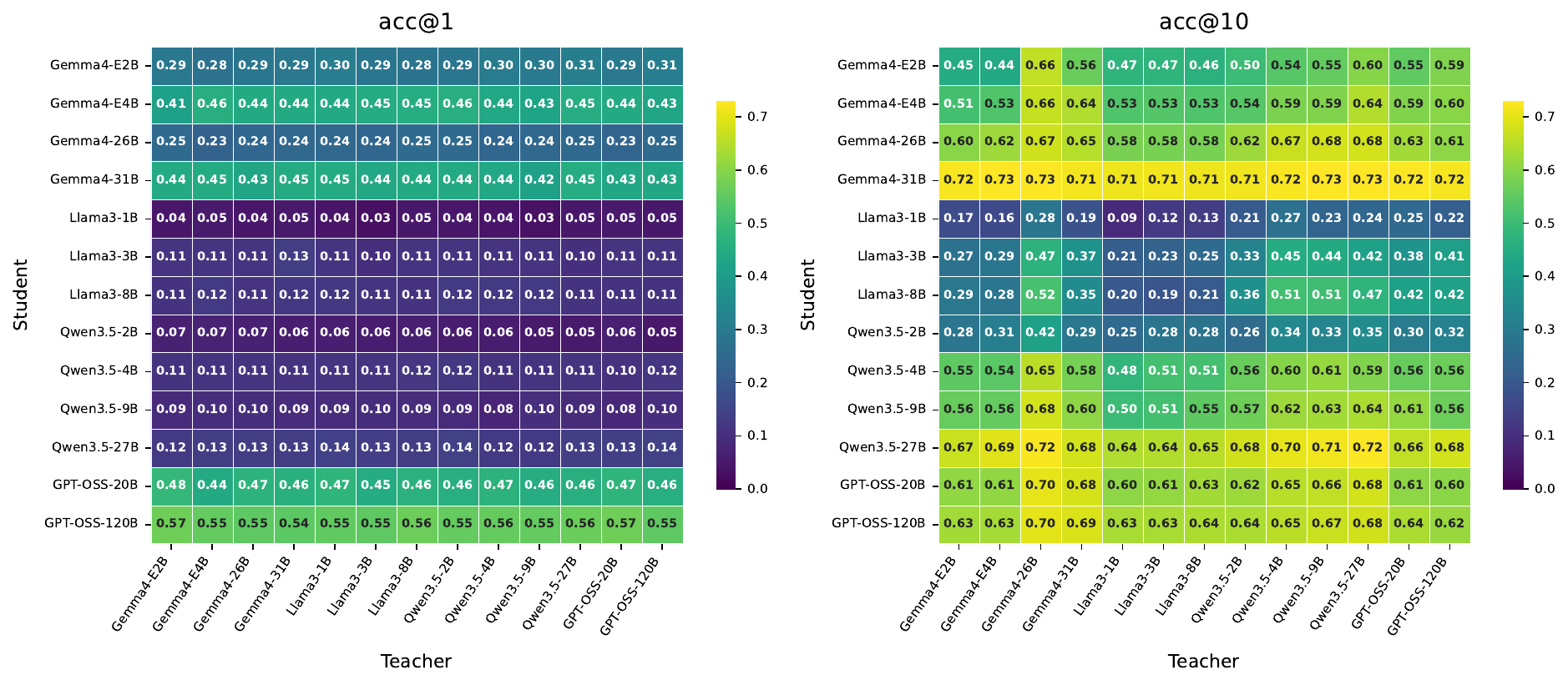}
\captionof{figure}{\textbf{Dense matrices separate zero-shot ability from feedback-mediated performance.}
Rows are students and columns are teachers for Omni-MATH. The left panel reports acc@1 before feedback, so values are constant across each student row; the right panel reports acc@10 after up to ten attempts with teacher feedback. Comparing the panels shows that feedback changes final performance substantially, some brighter columns highlight strong teachers, with the strongest structure remaining row-wise, indicating that the student model is a major driver of interactive outcomes.
}
\label{fig:omni-math-accuracy-matrices-main}
\end{center}

\paragraph{Models considered.}
To evaluate the performance of both teacher and student, we choose from thirteen models from four open-weight families, spanning a parameter range of 1B to 120B, both dense and Mixture-of-Experts (MoE) architectures, and models with and without dedicated reasoning modes. These are series of Gemma4-\{E2B, E4B, 26B-A4B, 31B\} \citep{team2026gemma}; the collection of Llama-3.x-\{1B, 3B, 8B\} \citep{dubey2024llama}; the family of Qwen3.5-\{2B, 4B, 9B, 27B\} \citep{qwen2026technical}; and the GPT-OSS models: GPTOSS-\{20B, 120B\} \citep{openai2025gptoss}.

\begin{figure}[t]
\centering
\includegraphics[width=0.85\linewidth]{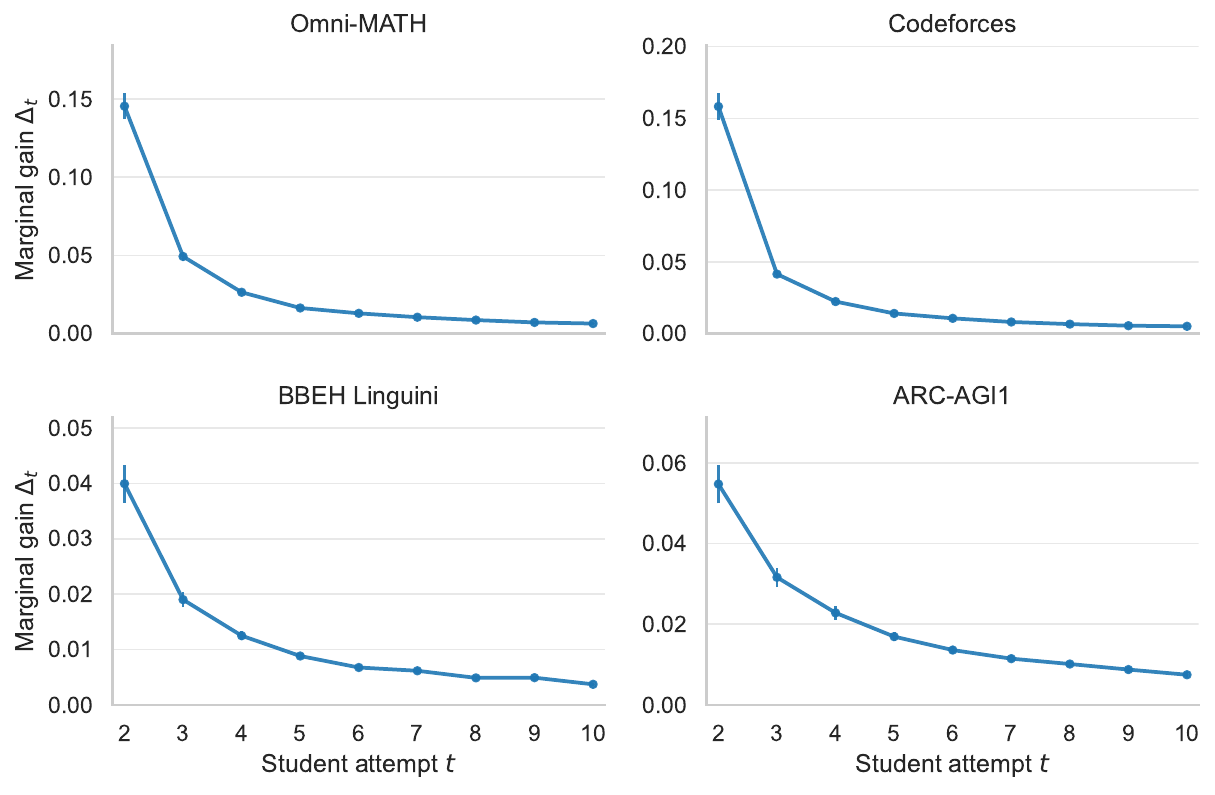}
\caption{\textbf{Most feedback gains arrive in the first few turns.}
Each panel reports the marginal improvement from one attempt to the next, averaged over dense-matrix student-teacher pairs. The first feedback-conditioned attempt produces the largest gain, and later turns contribute progressively smaller improvements. This front-loaded pattern suggests that short feedback trajectories capture most of the benefit, while longer rollouts are mainly useful for harder cases with slower recovery.}
\label{fig:marginal-gain-by-turn}
\end{figure}

\paragraph{Ablation tier.} \alicja{I'd move this to the appendix}
To further deepen our research, we conducted ablation studies to isolate the effect of different parameters on multi-turn performance. Due to the large search space and computational limitations, we focused solely on the Gemma-4 family, resulting in a $4 \times 4$ student-teacher matrix for each ablation. This provides us with a relatively exhaustive set of interactions between smaller and larger models. Because of the conversation setup, we test whether access to information from previous turns yields any benefits. For this purpose, we vary the maximum visible history length in $\{1,2,3,4,5\}$. Additionally, we test how the number of turns, also called the interaction budget, affects the final accuracy. We evaluate it for $K \in \{1,5,10\}$. Since the conversation progresses sequentially, we can generate the full sequence for $K=10$ and subsequently extract the states for $K=5$ and $K=1$ during post-processing. Finally, to decompose the gains from additional test-time compute, we added a self-refinement baseline for each model. This approach maintains the student's history but does not receive feedback from the teacher.

\paragraph{Metrics.}
To evaluate the different configurations and teacher-student pairs, we measure the accuracy at various steps of the interaction trajectory, denoted as $\text{acc}@K$. For a more comprehensive analysis, we introduce additional metrics: raw gain ($\text{acc}@K - \text{acc}@1$), normalized gain ($\frac{\text{acc}@K - \text{acc}@1}{1 - \text{acc}@1}$), and the Area Under the Curve computed as the mean cumulative accuracy over all turns ($\frac{1}{T} \sum_{k=1}^{T} \text{acc}@k$). Both $\text{acc}@K$ and AUC serve as primary measures of the system's overall problem-solving capability, additionally reflecting how quickly the student improves across turns. Raw and normalized gain instead isolate how much of that improvement is attributable to the teacher's feedback and iterative reasoning, independent of the student's starting level. Notably, normalized gain provides a saturation-adjusted comparison across different model scales by accounting for benchmark saturation, ensuring that larger models are not penalized for possessing a higher initial baseline accuracy ($\text{acc}@1$).

\begin{table}[t]
\centering
\caption{\textbf{Disentangling interaction, feedback, and feedback quality.}
Self-refinement measures improvement from additional attempts without external feedback;
self-feedback adds feedback generated by the same model; and best feedback selects the strongest available teacher for each student.
Cells report mean $\acc@10$ with mean $\gain@10$ in parentheses, in percentage points, averaged over students with all three conditions in the same environment.
Full per-student results are reported in \cref{tab:feedback-control-full}.}
\label{tab:feedback-control-averages}
{\small
\setlength{\tabcolsep}{4pt}
\begin{tabular}{@{}lccc@{}}
\toprule
Environment & Self-refinement & Self-feedback & Best feedback \\
\midrule
Omni-MATH & 43.9 (+20.7) & 48.6 (+24.7) & 60.5 (+36.9) \\
Codeforces & 52.8 (+17.8) & 57.0 (+23.8) & 68.7 (+35.1) \\
BBEH Linguini & 12.0 (+7.6) & 12.6 (+9.1) & 21.6 (+17.6) \\
ARC-AGI1 & 18.2 (+11.6) & 26.9 (+17.1) & 33.2 (+23.1) \\
\bottomrule
\end{tabular}
\par}
\vspace{-1.0em}
\end{table}

\section{What matters for improvement from feedback?}
\label{sec:results}

In this section we present our experimental results. We organize them around
the following six empirical questions, which we answer through research
findings (RFs):

\noindent
\begin{minipage}[t]{0.48\textwidth}
\vspace{0pt}
\begin{enumerate}[leftmargin=*]
    \item How much of multi-turn improvement comes from feedback rather than
    repeated attempts? (RF1)
    \item Can a student that is weaker on the first attempt become stronger
    after receiving feedback? (RF2)
    \item Is interactive performance driven primarily by the student receiving
    feedback or by the teacher generating it? (RF3)
    \item Is a teacher's single-turn task performance a good predictor of its
    competence as an interactive teacher? (RF4)
    \item How much improvement do longer interaction histories buy, given their
    cost and context-length tradeoffs? (RF5)
    \item When does teacher access to privileged information improve interactive outcomes? (RF6)
\end{enumerate}
\end{minipage}%
\hfill
\begin{minipage}[t]{0.5\textwidth}
\vspace{0pt}
\centering
\includegraphics[width=\linewidth]{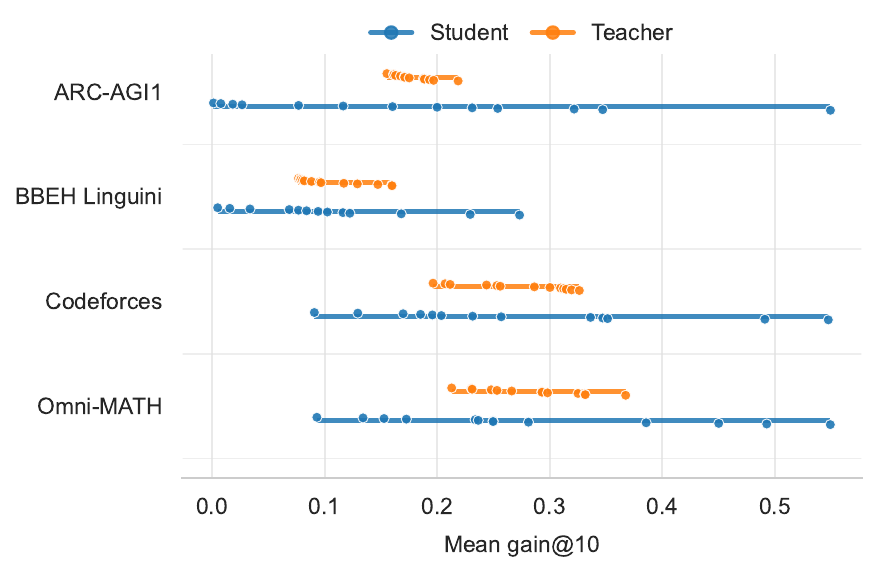}
\captionsetup{hypcap=false}
\captionof{figure}{\textbf{Student identity explains more gain variation than teacher identity.}
Points show role means at K = 10, averaging over the opposite role; segments show the min-to-max spread. Across environments, student means span a much wider range than teacher means.}
\label{fig:gain-by-student-teacher}
\end{minipage}

\subsection{Feedback-specific gains and interaction budget}

\paragraph{RF1: Feedback-specific gains are limited unless the teacher provides substantially better guidance than generic retry or self-feedback.}
The feedback-control comparison in \cref{tab:feedback-control-averages} shows
that multi-turn improvement is not, by itself, evidence that a student benefits from
feedback. Averaged over matched students, self-feedback is inconsistent relative to
the self-refinement baseline: it changes $\acc@10$ by $-1.2$ to $+8.7$ points
and $\gain@10$ by only $+0.1$ to $+6.0$ points across environments. The largest
self-feedback lift appears on ARC-AGI1, while BBEH Linguini slightly favors
unguided self-refinement in final accuracy.
Thus, simply adding a feedback message, even one generated by the same model,
does not guarantee feedback-specific improvement.
In contrast, selecting the best available teacher for each student adds
$9.2$ to $16.6$ points in $\acc@10$ and $9.9$ to $17.3$ points in $\gain@10$
over self-refinement. This suggests that gains depend not just on the presence
of a feedback message, but on whether the feedback provides corrective
information that the student can use.

\paragraph{Interaction budget also matters.}
The marginal-gain curves in \Cref{fig:marginal-gain-by-turn} show that the benefits of additional turns are front-loaded. Attempt 2, the first with feedback, accounts for $30.5\%$--$58.1\%$ of total $\gain@10$,
and attempts through $K=5$ recover $70.8\%$--$86.8\%$.
Later turns can still matter for hard long-horizon
cases, especially ARC-AGI1, but their average marginal return is small. 
For data collection, this favors shorter trajectories with informative feedback
over long rollouts whose gains mostly come from repeated attempts.

\subsection{Dense model matrix: separating student and teacher effects}
We next analyze the dense model matrix described in
\cref{sec:experimental-setup}: each model is evaluated both as a student and as
a teacher across the four environments, with episodes run for up to $K=10$
student attempts. This cross-product lets us ask whether interactive outcomes
are explained more by the model receiving feedback or by the model generating
it. \Cref{fig:omni-math-accuracy-matrices-main} shows a concrete slice of this
matrix for Omni-MATH, while the full dense matrices for all environments and
metrics are reported in \cref{app:model-matrix-figures}.

\begin{figure}[t]
\centering
\includegraphics[width=0.85\linewidth]{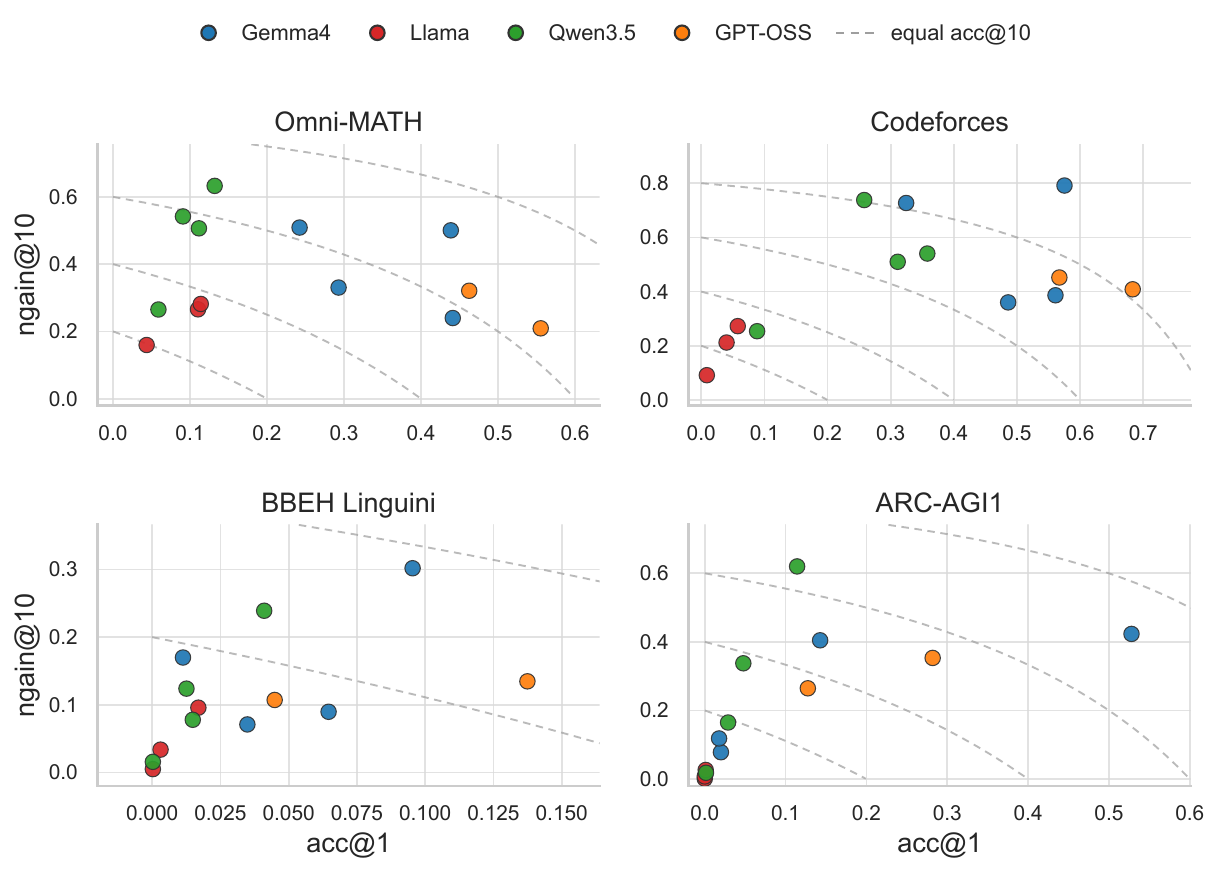}
\caption{\textbf{First-turn accuracy and feedback uptake are separate capabilities.}
Each panel shows one environment, and each point is a student model averaged
over teachers.  Color denotes model family. The y-axis reports
$\ngain@10=(\acc@10-\acc@1)/(1-\acc@1)$, the fraction of initially unsolved
problems recovered by the interaction. Dashed curves mark equal final
$\acc@10$ values. }
\label{fig:student-acc1-vs-gain}
\end{figure}

\paragraph{RF2: First-turn accuracy and feedback adherence are different
capabilities.}
A high initial solve rate does not imply that a model makes the best use of
feedback. 
\Cref{fig:student-acc1-vs-gain} shows that this distinction is not
a one-off comparison: students with similar $\acc@1$ can have very different
$\ngain$. A qualitative example can be seen in \ref{subsec:student_teacher_example}. It showcases how teacher and student cooperate to get to correct solution. A concrete comparison is given in
\cref{tab:single-turn-vs-plasticity} of \cref{app:plasticity-table}:
GPT-OSS-120B starts higher on Omni-MATH, Codeforces, and BBEH Linguini, but
Gemma4-31B usually has much larger $\gain$ and reaches the better final outcome.
ARC-AGI1 is the complementary case: GPT-OSS-120B has slightly larger $\gain$,
but Gemma4-31B starts much higher and remains the best final student. Thus
$\acc@1$ measures what the student can solve immediately, while $\ngain$ measures how efficiently it converts feedback rounds into recovery opportunities.

\paragraph{RF3: Interactive performance is mostly student-dependent.}
We use the dense matrix to marginalize over roles: for each student, we average
performance over teachers; for each teacher, we average performance over
students.  \Cref{fig:gain-by-student-teacher} shows the resulting role means
for $\gain$ at $K=10$.  Across all environments, the spread across students is much
larger than the spread across teachers, indicating that the model receiving 
feedback explains more of the interactive gain than the model generating it.
The appendix fixed-effect analysis in \cref{tab:main-dense-matrix-summary}
supports the same conclusion. Student fixed effects
explain $77.1\%$--$96.5\%$ of pair-level $\gain$ variation, while teacher fixed
effects explain only $1.3\%$--$12.4\%$. Adding teacher identity after student
identity gives a small incremental improvement, at most $0.124$ $R^2$.
This does not mean that teacher identity is irrelevant.  For a fixed student,
the gap between its best and worst teacher can be substantial, as shown by the
appendix teacher-range analysis in \cref{tab:average-teacher-range}.  The dense-matrix
results suggest that teacher effects are smaller, but can become
meaningful once we condition on a particular student.

\paragraph{RF4: Teacher task competence only partially predicts interactive teacher competence.}
We use each model's $\acc@1$ as a student as a noisy proxy for single-turn task
competence, and ask whether this ranking predicts its performance as a teacher
for a fixed student. If models that solve more problems on the first turn were
always stronger teachers, teacher rank and interactive $\auc$ would be
monotonic within each student.
\Cref{tab:teacher-monotonicity} shows that this monotonicity only partially
holds. Codeforces has the strongest relation, with mean Spearman
$\rho=0.56\pm0.02$, but the other environments are weaker
($0.21\pm0.06$ to $0.39\pm0.06$).
Inversions are also common: in $28.9\%\pm1.1\%$--$41.4\%\pm1.9\%$ of comparable
teacher pairs, the lower-$\acc@1$ teacher produces higher interactive $\auc$.
This suggests that single-turn solving ability and teaching ability are related
but not equivalent. A stronger solver may be more likely to know the correction,
but interactive performance also depends on diagnosing the student's specific
error, phrasing actionable feedback, and matching the student's ability to use
it.

\subsection{Gemma4 ablations: interaction history and privileged teacher information}

The dense matrix uses a relatively cheap and scalable protocol: short visible
history and no ground-truth answer or solution access for the teacher. We use
this as the main setting because both alternatives create tradeoffs. Longer
histories may help models diagnose repeated failures, but they increase cost and can run into context limits. Privileged information may make feedback
more accurate, but answers or solutions are not always available, may be costly
to obtain, and can leak into the teacher's feedback. The Gemma4 ablations
measure how much performance these more expensive sources of information buy.

\begin{figure}[t]
\centering
\includegraphics[width=\linewidth]{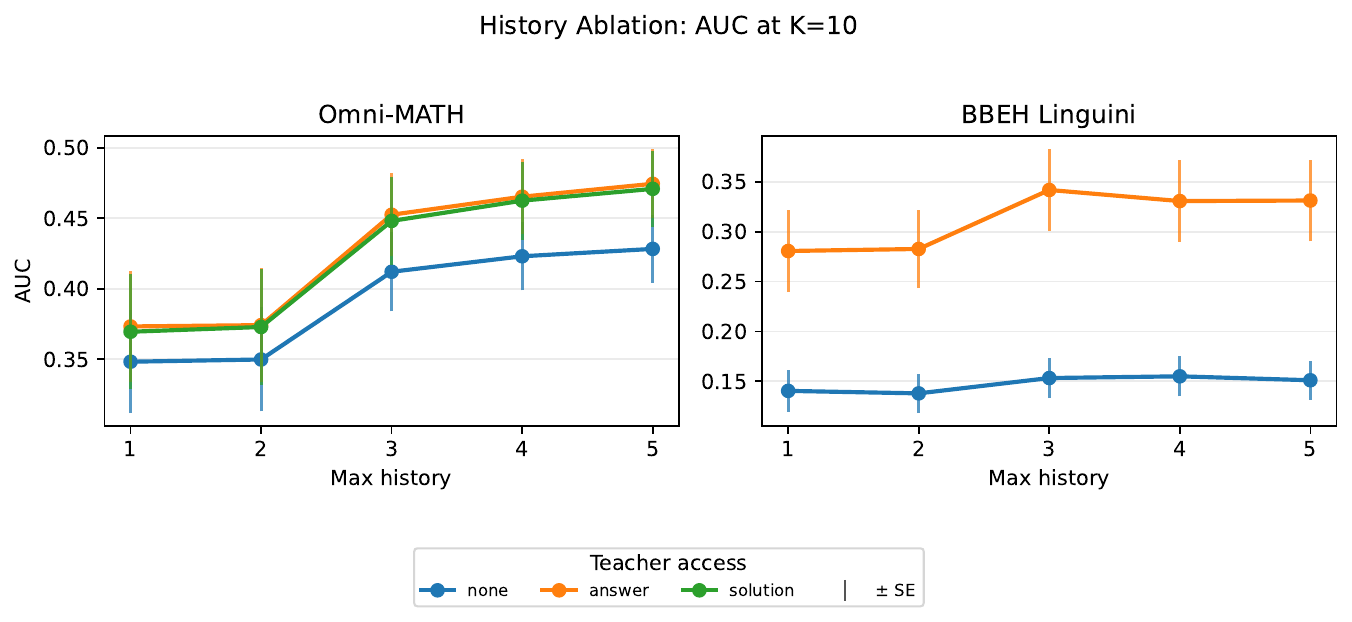}
\caption{\textbf{History and privileged teacher information help, but for different reasons.} 
Each panel reports $auc$ at $K=10$ as we vary the maximum number of previous interaction turns retained in the student and teacher prompts for Gemma4 student-teacher pairs, with line color indicating teacher reference access. 
}
\label{fig:gemma4-history-access-ablation-main}
\end{figure}

\paragraph{RF5: Longer visible history is a capability-dependent tradeoff.}
Longer histories expose repeated failure modes to the teacher and give the
student access to earlier corrections. This information is not free: every
additional visible turn increases cost and makes context overflow more likely.
It is also not automatically useful, the models must be able to identify which
parts of the history matter and use them to revise the next attempt.
The max-history ablation shows little change from history 1 to 2, followed by a
larger jump at history 3. Averaged over teacher-access settings and Gemma4 pair
cells, moving from history 1 to 5 increases $\auc$ by $0.094$ on Omni-MATH. In
the full grid, Codeforces shows the largest average history benefit, with
$\auc$ increasing by $0.123$. BBEH Linguini shows a smaller but positive gain,
rising from $0.210$ at history 1 to $0.248$ at history 3. These averages suggest
that history can help, but the benefit should be interpreted as conditional on
model capacity rather than as a free improvement from longer context.

This capability dependence is visible when we split by teacher-student strength. \mg{as i remember max history is an attribute to both student and teacher. Maybe looking into pairs of teacher-student weak-weak, strong-strong would give different results. Probably history for weaker models hurts their performance and we do not show it here. }
In the matched history-5 minus history-1 diagnostic
(\cref{fig:gemma4-history-lift-by-teacher-strength}), Codeforces gains about $0.21$ $\auc$ for
pairs where the teacher's single-turn $\acc@1$ exceeds the student's, compared
with about $0.05$ for the remaining pairs. This is consistent with stronger
teachers using history to diagnose repeated failures, although the ablation
still does not isolate teacher adaptation from student-side conditioning.
\todo{update the plots with ARC-AGI ablations and mention them here}
\todo{it would be nice to quantify the cost of improvement. You can improve but you need to pay X more...}

\paragraph{RF6: Privileged information helps selectively, not uniformly.}
We ask whether giving the teacher additional task-side information reliably
improves the interaction. The answer is task-dependent. At history 5, BBEH
Linguini shows the largest effect: answer access increases $\auc$ from $0.151$
to $0.331$ and $\acc@10$ from $0.195$ to $0.463$. This suggests that privileged
information is most useful when the teacher would otherwise struggle to infer
the target correction. This large BBEH separation is visible in
\cref{fig:gemma4-history-access-ablation-main}
The effect is much smaller elsewhere. On Omni-MATH, answer access improves
$\auc$ from $0.428$ to $0.475$, while full solution access reaches only $0.471$.
On Codeforces, where the teacher already receives structured execution feedback,
adding the reference solution changes $\auc$ only from $0.547$ to $0.556$.
ARC-AGI1 shows a smaller but measurable answer-access effect: $\auc$ increases
from $0.303$ to $0.328$, while $\acc@10$ increases from $0.362$ to $0.408$.
Thus, additional teacher-side information is not automatically useful; it helps
when it resolves a real teacher uncertainty and can be converted into actionable
feedback. The teacher-level lift diagnostic in
\cref{fig:gemma4-teacher-information-lift-by-teacher} shows the same
non-uniformity across teachers and environments, and the full access summary is
shown in \cref{fig:app-gemma4-teacher-access-ablation-full}.
Because answers and solutions are not always available and may leak into the
feedback, we treat these conditions as upper-bound ablations rather than the
default scalable protocol. To quantify how privileged information affects
solution leakage from teacher to student, we conduct a leakage audit in
\cref{fig:leak-rate-absolute} of \cref{app:leakage-audit}.
\todo{this section needs a leakage audit}

\section{Discussion}
\label{sec:conclusion}


\paragraph{Summary}
Our findings have practical consequences for evaluating and building feedback-based agents. First, repeated-attempt and self-feedback controls should be included whenever multi-turn gains are reported; otherwise, extra inference compute can be mistaken for feedback use. Second, teacher selection should not be based only on single-turn task accuracy. The teacher monotonicity analysis shows that stronger solvers are not always stronger teachers for a particular student. Third, longer histories and privileged teacher information should be treated as targeted interventions rather than default improvements: they help when they resolve a real diagnostic bottleneck, but they also increase cost, context pressure and leakage risk.

\paragraph{Limitations}
Our study has several limitations. First, the current ARC-AGI1 analysis does not
yet include the same ablations as the other environments.
Although we provided preliminary qualitative analysis of the feedback trajectories, 
conducting a broader analysis in the future will be necessary to achieve more conclusive results. These analyses
are important for understanding whether improvements come from actionable
feedback rather than formatting artifacts, repeated sampling, or direct leakage
from the teacher context.
Second, our experiments focus on a zero-shot setting with short-horizon
verifiable tasks and open-weight models. This makes the protocol controlled and
reproducible, but it leaves open how the conclusions transfer to models trained
to use feedback, longer-horizon interactive domains, and closed-source systems.

\paragraph{Future Work}
A natural direction for future work is to connect this diagnostic protocol to
post-training. In supervised fine-tuning or reinforcement-learning settings,
feedback trajectories could be used not only to measure whether a model improves
within an episode, but also to train models that become better feedback users
over time. Extending the study to longer-horizon domains such as computer usage
would test whether the same student-side bottleneck appears when feedback must
be integrated across many actions, persistent state, and delayed consequences.
Finally, comparing open-weight and closed-source models would clarify which
findings reflect general properties of current language models and which depend
on the model families studied here.

\bibliographystyle{plainnat}
\bibliography{references}

\appendix
\section{Experimental Setup Details}
\label{app:experimental-details}

\subsection{Protocol Definition}
\label{app:protocol-definition}

Our evaluation protocol follows the didactic feedback loop of \citet{klissarov2026interactivefeedback}, but uses it only as a zero-shot evaluation environment. A single episode contains a task instance, a student model, a teacher model, and a maximum number of student attempts. The first student message contains the task but no teacher feedback. After each student attempt, the environment evaluates correctness. Correct attempts terminate the episode immediately. Incorrect attempts trigger a teacher turn unless the attempt budget has already been exhausted.

Let $x$ be the public task prompt, $z$ be optional privileged task information, and $h_t$ be the prompt-visible interaction history before the $t$-th student attempt. The student generates
\[
  a_t \sim \pi_S(\cdot \mid x, h_t, f_{t-1}),
\]
where $f_{t-1}$ is empty on the first turn. The verifier returns $c_t=1$ if $a_t$ is correct. If $c_t=0$ and $t<K_{\max}$, the teacher generates
\[
  f_t \sim \pi_T(\cdot \mid x, a_t, h_t, z_T),
\]
where $z_T$ is determined by the teacher-information condition. In the dense model matrix, $z_T$ excludes the final answer and full solution for math and BBEH, and excludes the trusted reference solution for code. In code, the teacher can still see execution feedback unless explicitly ablated.

The teacher is always instructed to provide diagnostic feedback rather than the final answer. In the main prompt style, the teacher may reason in \texttt{<think>} tags, but the feedback passed to the student is parsed from the final \texttt{<feedback>} block. If parsing fails, the raw teacher text is passed through.

\paragraph{History truncation.}
The implementation maintains separate student-side and teacher-side chat histories. The parameter \texttt{max\_history} controls how many completed interaction turns are kept in the prompt. When the history window is full, older refinement turns are removed before adding the next user message. The original user prompt containing the task description is preserved, so every later student or teacher call still has the problem context and the chat remains a natural continuation of the same task. 

\subsection{Run Configuration}
\label{app:run-configuration}

\begin{table}[h]
\centering
\small
\begin{tabular}{llll}
\toprule
Environment & Episodes & Teacher privileged information & Extra teacher context \\
\midrule
Omni-MATH & 512 & none & none \\
Codeforces & 512 & none & structured execution feedback \\
BBEH Linguini & 300 & none & none \\
ARC-AGI1 & 400 & none & structured ARC feedback \\
\bottomrule
\end{tabular}
\caption{\textbf{Current dense-matrix run settings.}}
\label{tab:matrix-run-settings}
\end{table}

The current dense-matrix runs use the following common settings:
\begin{itemize}
  \item Agent: \texttt{hierarchical\_separate}, with separate vLLM actor instances for student and teacher.
  \item Interaction budget: $K_{\max}=10$ student attempts.
  \item History: \texttt{max\_history=1}.
  \item Student sampling: temperature $0.7$, top-$p=0.95$, max tokens $8192$.
  \item Teacher sampling: temperature $1.0$, top-$p=0.95$, max tokens $8192$.
  \item Context length: $32768$ tokens.
  \item Max output tokens: $8192$ tokens.
  \item Number of workers: 16 evaluation workers.
\end{itemize}

ARC-AGI uses max output tokens $16000$ for both student and teacher because the task JSON and output grids can be long.

\subsection{Model Set}
\label{app:model-set}

The dense matrix uses four model families totaling to thirteen model identifiers:
\begin{itemize}
  \item Gemma4: 
    \begin{itemize}
        \item \texttt{gemma-4-E2B-it}, 
        \item \texttt{gemma-4-E4B-it}, 
        \item \texttt{gemma-4-26B-A4B-it}, 
        \item \texttt{gemma-4-31B-it},
    \end{itemize}
    
  \item Qwen3.5: 
    \begin{itemize}
        \item \texttt{Qwen3.5-2B},
        \item \texttt{Qwen3.5-4B},
        \item \texttt{Qwen3.5-9B},
        \item \texttt{Qwen3.5-27B},
    \end{itemize}
       
  \item Llama: 
      \begin{itemize}
        \item \texttt{Llama-3.2-1B-Instruct},
        \item \texttt{Llama-3.2-3B-Instruct},
        \item \texttt{Llama-3.1-8B-Instruct},
      \end{itemize}
    
  \item GPT-OSS: 
    \begin{itemize}
        \item \texttt{gpt-oss-20b},
        \item \texttt{gpt-oss-120b}.
    \end{itemize}

\end{itemize}
Each model appears once as the student and once as the teacher, for 169 pairs per dense-matrix environment.

\subsection{Datasets and Verifiers}
\label{app:datasets-verifiers}

\paragraph{Omni-MATH.}
Each example contains a problem, final answer, ground-truth solution, and source. The student is asked to end with one boxed final answer. Correctness is evaluated by two math answer checkers: a PRIME-style normalization and symbolic equivalence checker, followed by a \texttt{math\_verify}-based checker. The environment returns reward 1 if one checker accepts the answer.

\paragraph{Codeforces.}
Each example contains a prompt, optional starter code, trusted reference solution, test specification, source, and difficulty metadata. The student returns a complete Python solution in a fenced code block. Candidate code is run in a subprocess sandbox on at most 16 test cases. A task is solved only when all tests pass. Current timeouts use dataset-aware scaling: per-case timeout defaults to 10 seconds, is multiplied by 10 when a dataset time limit is available, and is clamped between 5 and 20 seconds; the full-task timeout is derived from the case timeout and number of cases.

\paragraph{BBEH (Linguini).}
Each example contains a task name, problem text, reference answer, source, and deterministic problem id derived from the task, problem, and answer. We filter the BBEH dataset to the Linguini task. Correctness is evaluated by lowercasing and normalizing the prediction and reference, stripping simple LaTeX wrappers, handling common answer prefixes, and applying a small set of fuzzy matches for parenthesized choices, numeric strings, quotes, brackets, and trailing question marks. The environment returns reward 1 if the normalized prediction matches the reference under this evaluator.

\paragraph{ARC-AGI1.}
Each example contains a task id, training input/output grid pairs, test input grids, and hidden target output grids used only by the verifier. The student must return a JSON object, using the schema \texttt{\{"outputs": [[[...]], ...]\}}. Correctness is evaluated by parsing the JSON, validating that each predicted output is a well-formed ARC grid with the expected number of test outputs, and comparing the predicted grids to the hidden target grids. The environment returns reward 1 only when all test outputs match exactly. Structured feedback reports parsing failures, shape errors, or grid mismatch diagnostics, truncated to 2000 characters.

\subsection{Teacher Information Conditions}
\label{app:teacher-information-conditions}

We use the following teacher reference modes:
\begin{itemize}
  \item \texttt{none}: the teacher receives no ground-truth answer or solution.
  \item \texttt{answer}: the teacher receives only the final answer or reference target.
  \item \texttt{solution}: the teacher receives the full ground-truth solution or trusted reference solution.
\end{itemize}
For code, reference mode is crossed with execution-context mode:
\begin{itemize}
  \item \texttt{structured}: number of tests passed, pass rate, timeout status, and first-failure details when available.
  \item \texttt{feedback}: the environment's compact textual execution feedback.
  \item \texttt{traceback}: only stderr or traceback text when available.
  \item \texttt{none}: no execution-side signal beyond the student code and problem prompt.
\end{itemize}

\subsection{Leakage audit}
\label{app:leakage-audit}
For the purpose of evaluating we have chosen two out of four environments, specifically Omni-MATH and BBEH Linguini. To assess whether the models leak information, we prompted an external large language model judge with a triplet containing the (solution, student answer, teacher response).  The judge model was instructed to return True if the teacher leaked confidential information that trivially helped the student solve the problem. Fixing the student model as Gemma 4 26B, we iterated across all teacher models to quantify the extent of information leakage. We calculated the leak rate as the ratio of identified leaks to the total number of judge responses.

\begin{figure}[ht]
\centering
\includegraphics[width=1.0\linewidth]{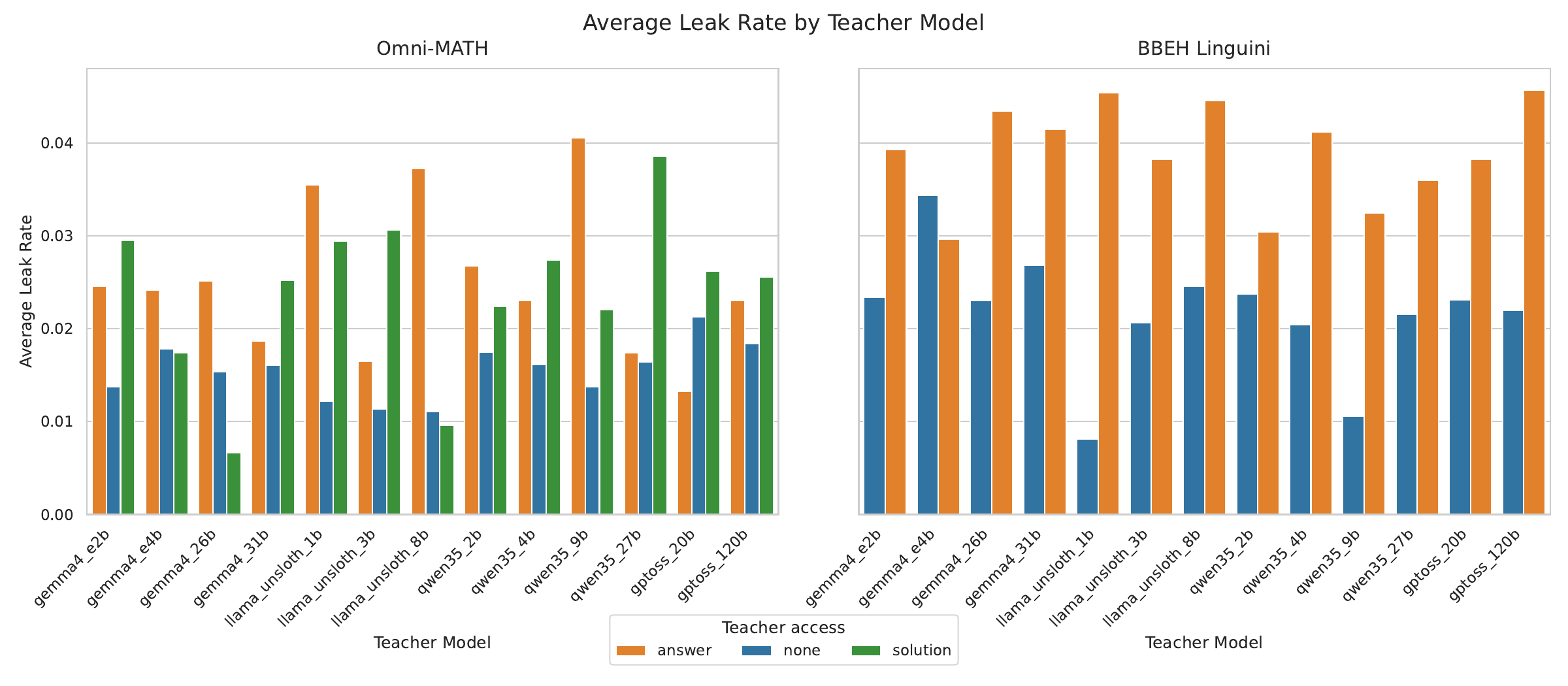}
\caption{Absolute leak rates across various teacher models and teacher access types, evaluated with a fixed student model (Gemma 4 26B).}
\label{fig:leak-rate-absolute}
\end{figure}

We can see in \Cref{fig:leak-rate-absolute} that no more than 5\% of teacher answers contained a leak information. Consistently no additional teacher access scores lower leak rates compared to both answer access and solution access. The differences mostly vary in smaller models.

\subsection{Metrics from Episode Logs}
\label{app:metrics-from-logs}

Each completed episode log contains at least \texttt{solved}, \texttt{attempts\_used}, \texttt{max\_turns}, and, when available, \texttt{problem\_id}. For each cutoff $k$, the postprocessing code marks an episode as solved by $k$ if \texttt{solved=1} and \texttt{attempts\_used <= k}. If multiple episodes share a problem id, the code averages within problem first and then averages across problems. Otherwise each episode is treated as its own problem.

The main reported metrics are:
\begin{equation}
    \acc@k = \frac{1}{N}\sum_i \mathbf{1}[\text{problem } i \text{ solved by } k]
\end{equation}

\begin{equation}
    \gain@K = \acc@K-\acc@1,
\end{equation}

\begin{equation}
    \ngain@K =
\begin{cases}
(\acc@K-\acc@1)/(1-\acc@1), & \acc@1 < 1\\
0, & \acc@1 = 1
\end{cases}
\end{equation}

\begin{equation}
    \auc = \frac{1}{K_{\max}}\sum_{k=1}^{K_{\max}}\acc@k
\end{equation}

We also report $\acc@2-\acc@1$ because a very large first refinement jump can indicate a formatting or first-turn protocol artifact rather than steady feedback use.

\subsection{Prompt Templates}
\label{app:prompt-templates}

This section records the prompt templates used by the current implementation. Braced names such as \texttt{\{problem\}} denote task-specific fields inserted at runtime.

\paragraph{Shared teacher system prompt.}
\begin{verbatim}
You are a careful tutor. Read the student's latest attempt, identify the most
important mistake, and give targeted feedback without revealing the final answer.
\end{verbatim}

\paragraph{Shared teacher output format.}

\begin{verbatim}
Pinpoint the step in which I am making a mistake.
Provide the most informative piece of information for me to succeed on the next
try, without telling me the final answer.
You may reason about the mistake first inside <think>...</think> tags.
Your final feedback message must appear exactly once inside
<feedback>...</feedback> tags.

Return only:
<feedback>
your feedback
</feedback>
\end{verbatim}

\paragraph{Omni-MATH student prompts.}
\begin{verbatim}
System:
You are a mathematician solving competition-level math problems.

Initial user prompt:
Problem:
{problem}

Solve the problem carefully.

You may reason in <think>...</think>. End with exactly one final line in this
format:
\boxed{your answer}

Refinement user prompt, feedback condition:
Feedback:
{teacher_feedback}

Revise your previous solution so it uses the feedback.

You may reason in <think>...</think>. End with exactly one final line in this
format:
\boxed{your answer}

Refinement user prompt, self-refinement/no-feedback condition:
Revise your previous solution and produce a corrected solution.

You may reason in <think>...</think>. End with exactly one final line in this
format:
\boxed{your answer}
\end{verbatim}

\paragraph{Omni-MATH teacher prompts.}
\begin{verbatim}
Initial teacher user prompt:
I am trying to solve this question:
{problem}

Here is my current attempt:
{previous_attempt}

{reference_context}

{teacher_output_rules}

Follow-up teacher user prompt:
Here is my current attempt:
{previous_attempt}

{teacher_output_rules}
\end{verbatim}
The reference context is one of:
\begin{verbatim}
You do not have access to the ground truth solution or final answer.

Here is the ground-truth final answer:
{answer}

Here is the ground-truth solution:
{solution}
\end{verbatim}

\subsection{Example of interaction between student and teacher}
\label{subsec:student_teacher_example}
This section provides an example of a complete dialogue between teacher gpt-oss 120B and student Gemma-4-31B. To make it more parsable outputs of student have been summarized. Teacher in this trajectory provided student with necessary refinement to complete the task.

\textbf{ Agent Interaction: Student-Teacher Correction}

\noindent\textbf{Student [Step 0] --- Reasoning \& Initial Hypothesis:}\\
Analyzes the input grids and identifies the bounding box of the active shape (color 4). Hypothesizes that the target output is a $3 \times 3$ grid of blocks forming a cross shape. Attempts to deduce the geometric transformations for each arm of the cross, but incorrectly formulates the policy: assuming the right arm is a horizontal reflection and the bottom arm is a vertical reflection, rather than pure rotations.

\vspace{0.75em}
\noindent\textbf{Student [Step 0] --- Action:}\\
Submits a $17 \times 17$ output grid. (\textit{Result: Incorrect, Reward: 0.0})

\vspace{0.75em}
\noindent\textbf{Teacher [Step 1] --- Feedback:}\\
``You correctly noticed that the output is built from four copies of the input's colored shape arranged in a cross, but the transformations you applied to the arms are not the ones used in the examples. In the training pairs, the top arm is the shape rotated $90^\circ$ clockwise, the left arm is the original shape, the right arm is the shape rotated $180^\circ$ (i.e., both axes flipped), and the bottom arm is the shape rotated $270^\circ$ clockwise (or equivalently a $90^\circ$ counter-clockwise rotation). 

For non-square shapes, the rotated arms swap height and width, so the top and bottom blocks have the dimensions of the input $W \times H$, while the left and right blocks keep the original $H \times W$. Because you used a horizontal flip... several cells are placed in the wrong positions. Re-examine the transformations: left = original, top = $90^\circ$ cw, right = $180^\circ$, bottom = $270^\circ$ cw, and assemble the output accordingly.''

\vspace{0.75em}
\noindent\textbf{Student [Step 1] --- Reasoning \& Correction:}\\
Integrates the teacher's feedback to update the transformation policy. Re-evaluates the active bounding box, explicitly noting it is non-square ($H=5, W=6$). Calculates the necessary dimensional swaps for the top and bottom arms ($W=5, H=6$). Methodically maps the pixel coordinates for $S_{90}$, $S_{180}$, and $S_{270}$ applied to the base shape $S$, and reconstructs the spatial layout of the final grid using the corrected rotational rules.

\vspace{0.75em}
\noindent\textbf{Student [Step 1] --- Action:}\\
Submits the corrected $17 \times 17$ output grid. (\textit{Result: Success, Reward: 1.0})

\section{Additional Results}

    The initial sections present detailed dense-matrix tables and performance visualizations for all model pairings across the tested environments, including Omni-MATH, Codeforces, BBEH Linguini, and ARC-AGI1. They do it for all types of our metrices we used: accuracy, cumulative accuracy, and performance gains.

    Subsequent sections focus on specific ablation studies and deeper mechanistic insights, In them we are using gemma 4 dense matrices.
   We isolate the impact of relative teacher strength, information access and maximum history, providing a thorough analysis of the variables that govern reasoning improvements in multi-turn feedback settings.
\label{app:additional-results}
\subsection{Supplementary Dense-Matrix Tables}

\label{app:dense-matrix-tables}

\begin{table}[t]
\centering
\caption{\textbf{Most pair-level gain variation is explained by the student.}
Metrics are averaged over student-teacher pair cells at $K=10$.  Student FE and
teacher FE are single-factor fixed-effect $R^2$ values for pair-level
$\gain=\acc@10-\acc@1$;
the last column is the incremental $R^2$ from adding teacher identity after
student identity.}
\label{tab:main-dense-matrix-summary}
\resizebox{\linewidth}{!}{%
\begin{tabular}{lrrrrrrr}
\toprule
Environment & $\acc@1$ & $\acc@10$ & $\gain$ & $\auc$
& Student FE $R^2$ & Teacher FE $R^2$ & $\Delta R^2$ teacher$|$student \\
\midrule
Omni-MATH & 0.238 & 0.521 & 0.283 & 0.452 & 0.858 & 0.093 & 0.093 \\
Codeforces & 0.332 & 0.604 & 0.272 & 0.544 & 0.827 & 0.097 & 0.097 \\
BBEH Linguini & 0.037 & 0.144 & 0.107 & 0.111 & 0.799 & 0.132 & 0.132 \\
ARC-AGI1 & 0.101 & 0.278 & 0.178 & 0.219 & 0.963 & 0.012 & 0.012 \\
\bottomrule
\end{tabular}
}
\vspace{-1.0em}
\end{table}

\begin{table}[t]
\centering
\caption{\textbf{Teacher effects are smaller globally but still consequential for a fixed student.}
For each student, we compute the gap between the best and worst teacher at
$K=10$, then average these within-student gaps over students.}
\label{tab:average-teacher-range}
\resizebox{\linewidth}{!}{%
\begin{tabular}{lrrrr}
\toprule
Environment &  Mean worst $\auc$ & Mean best $\auc$
& Mean $\Delta\auc$ & Mean $\Delta\acc@10$ \\
\midrule
Omni-MATH & 0.395 & 0.511 & 0.116 & 0.158 \\
Codeforces & 0.483 & 0.608 & 0.125 & 0.163 \\
BBEH Linguini & 0.079 & 0.163 & 0.085 & 0.117 \\
ARC-AGI1 & 0.191 & 0.256 & 0.065 & 0.090 \\
\bottomrule
\end{tabular}
}
\vspace{-1.0em}
\end{table}

\begin{center}
\centering
\captionof{table}{\textbf{Teacher task competence is only partially predictive of
interactive teacher competence.}  We rank teachers by empirical $\acc@1$ when
acting as a student in the same environment, using it as a noisy proxy for
single-turn task competence.  Spearman correlations are computed within each
fixed student between this teacher-rank proxy and interactive $\auc$.  Inversion
rate is the fraction of comparable teacher pairs where the lower-$\acc@1$
teacher gives higher interactive $\auc$.  Values are means over fixed-student
groups $\pm$ one standard error.}
\label{tab:teacher-monotonicity}
\begin{tabular}{lcc}
\toprule
Environment & Mean Spearman $\rho$ & Inversion rate \\
\midrule
Omni-MATH & $0.21\,{\color{gray}\scriptstyle \pm 0.06}$ & $41.4\%\,{\color{gray}\scriptstyle \pm 1.9\%}$ \\
Codeforces & $0.53\,{\color{gray}\scriptstyle \pm 0.04}$ & $29.9\%\,{\color{gray}\scriptstyle \pm 1.7\%}$ \\
BBEH Linguini & $0.44\,{\color{gray}\scriptstyle \pm 0.07}$ & $33.4\%\,{\color{gray}\scriptstyle \pm 2.5\%}$ \\
ARC-AGI1 & $0.33\,{\color{gray}\scriptstyle \pm 0.11}$ & $35.4\%\,{\color{gray}\scriptstyle \pm 4.0\%}$ \\
\bottomrule
\end{tabular}
\vspace{-1.0em}
\end{center}

\clearpage
\subsection{Supplementary Plasticity Table}
\label{app:plasticity-table}

\begin{center}
\centering
\captionof{table}{\textbf{High acc@1 can lose to better feedback adherence.}
The table compares GPT-OSS-120B and Gemma4-31B, averaging metrics over teachers.
GPT-OSS-120B often starts from a stronger first-turn baseline, but Gemma4-31B
achieves higher interactive $\auc$ across environments because it gains more from
feedback. This supports the claim that feedback adherence is distinct from initial
task competence. Grey terms report standard errors over teachers.}
\label{tab:single-turn-vs-plasticity}
\resizebox{\linewidth}{!}{%
\begin{tabular}{lrrrrrrrr}
\toprule
& \multicolumn{4}{c}{GPT-OSS-120B}
& \multicolumn{4}{c}{Gemma4-31B} \\
\cmidrule(lr){2-5}\cmidrule(lr){6-9}
Environment & $\acc@1$ & $\acc@10$ & $\gain$ & $\auc$
& $\acc@1$ & $\acc@10$ & $\gain$ & $\auc$ \\
\midrule
Omni-MATH & $0.555\,{\color{gray}\scriptstyle \pm 0.002}$ & $0.648\,{\color{gray}\scriptstyle \pm 0.007}$ & $0.093\,{\color{gray}\scriptstyle \pm 0.008}$ & $0.624\,{\color{gray}\scriptstyle \pm 0.006}$ & $0.439\,{\color{gray}\scriptstyle \pm 0.002}$ & $0.720\,{\color{gray}\scriptstyle \pm 0.002}$ & $0.281\,{\color{gray}\scriptstyle \pm 0.004}$ & $0.679\,{\color{gray}\scriptstyle \pm 0.002}$ \\
Codeforces & $0.683\,{\color{gray}\scriptstyle \pm 0.002}$ & $0.813\,{\color{gray}\scriptstyle \pm 0.009}$ & $0.129\,{\color{gray}\scriptstyle \pm 0.008}$ & $0.782\,{\color{gray}\scriptstyle \pm 0.008}$ & $0.575\,{\color{gray}\scriptstyle \pm 0.003}$ & $0.911\,{\color{gray}\scriptstyle \pm 0.003}$ & $0.336\,{\color{gray}\scriptstyle \pm 0.004}$ & $0.864\,{\color{gray}\scriptstyle \pm 0.003}$ \\
BBEH Linguini & $0.137\,{\color{gray}\scriptstyle \pm 0.005}$ & $0.254\,{\color{gray}\scriptstyle \pm 0.015}$ & $0.116\,{\color{gray}\scriptstyle \pm 0.013}$ & $0.213\,{\color{gray}\scriptstyle \pm 0.012}$ & $0.095\,{\color{gray}\scriptstyle \pm 0.002}$ & $0.368\,{\color{gray}\scriptstyle \pm 0.008}$ & $0.273\,{\color{gray}\scriptstyle \pm 0.008}$ & $0.303\,{\color{gray}\scriptstyle \pm 0.006}$ \\
ARC-AGI1 & $0.282\,{\color{gray}\scriptstyle \pm 0.003}$ & $0.536\,{\color{gray}\scriptstyle \pm 0.011}$ & $0.254\,{\color{gray}\scriptstyle \pm 0.010}$ & $0.453\,{\color{gray}\scriptstyle \pm 0.009}$ & $0.528\,{\color{gray}\scriptstyle \pm 0.003}$ & $0.728\,{\color{gray}\scriptstyle \pm 0.004}$ & $0.200\,{\color{gray}\scriptstyle \pm 0.005}$ & $0.670\,{\color{gray}\scriptstyle \pm 0.003}$ \\
\bottomrule
\end{tabular}
}
\end{center}

\clearpage
\subsection{Supplementary Model-Matrix Figures}
\label{app:model-matrix-figures}

\begin{figure}[ht]
\centering
\includegraphics[width=1.0\linewidth]{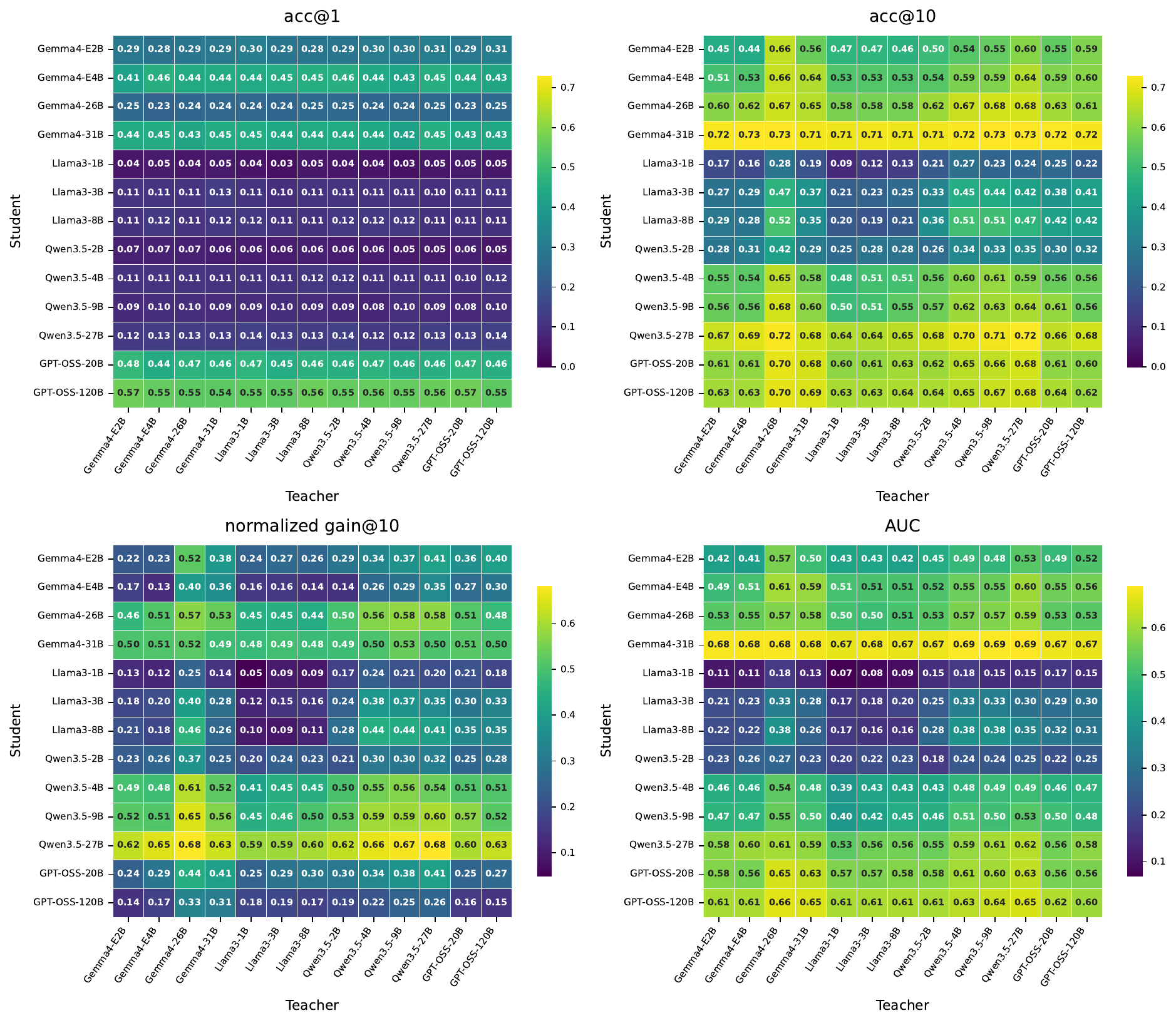}
\caption{\textbf{Omni-MATH dense model matrices}. Rows are students and columns are teachers.  We annotate all four metrics: $\acc@1$, $\acc@10$, $\gain$, and cumulative accuracy $\auc$.}
\label{fig:app-omni-math-matrix}
\end{figure}

\begin{figure}[ht]
\centering
\includegraphics[width=1.0\linewidth]{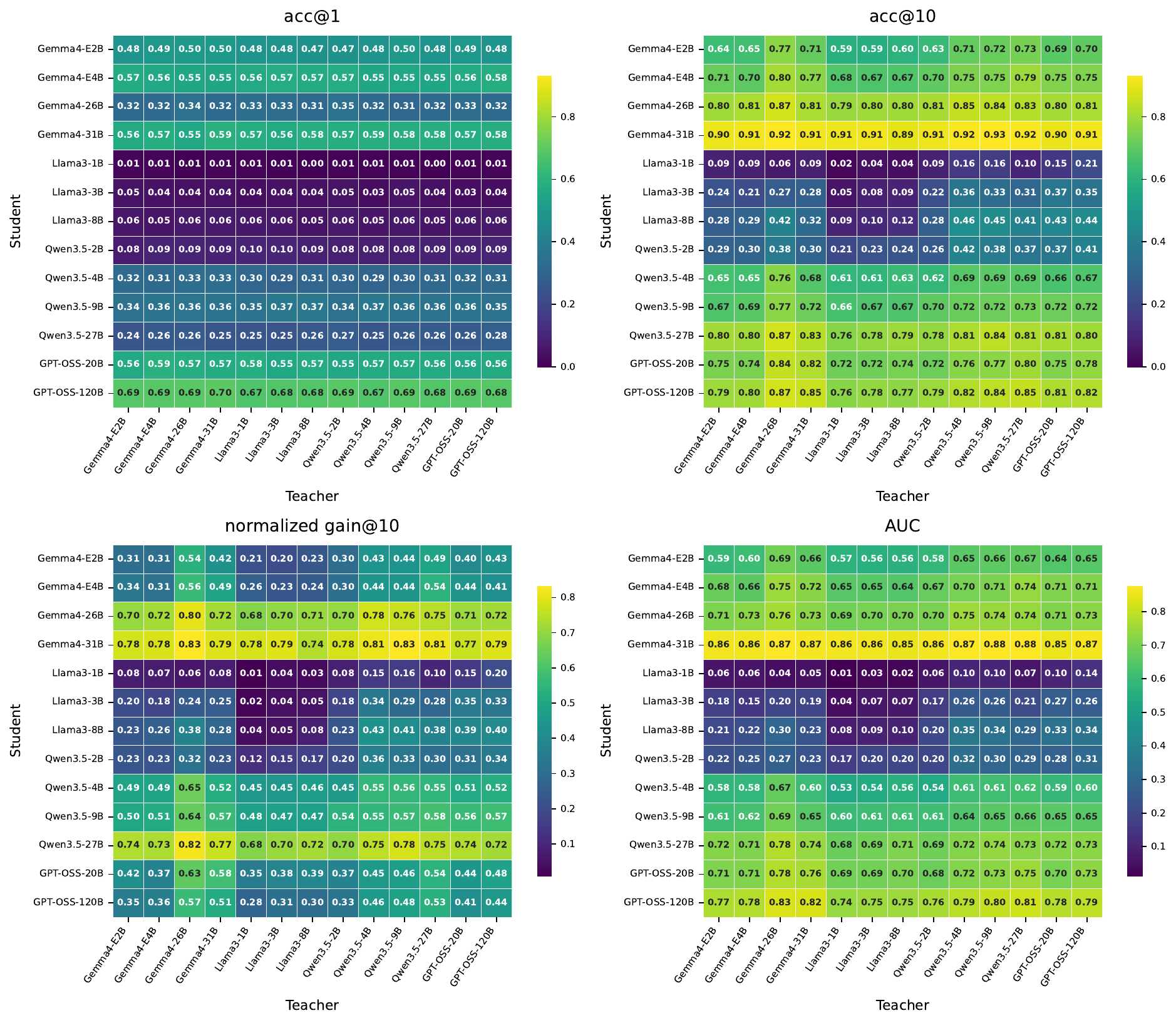}
\caption{\textbf{Codeforces dense model matrices}. Rows are students and columns are teachers.  We annotate all four metrics: $\acc@1$, $\acc@10$, $\gain$, and cumulative accuracy $\auc$.}
\label{fig:app-codeforces-matrix}
\end{figure}

\begin{figure}[ht]
\centering
\includegraphics[width=1.0\linewidth]{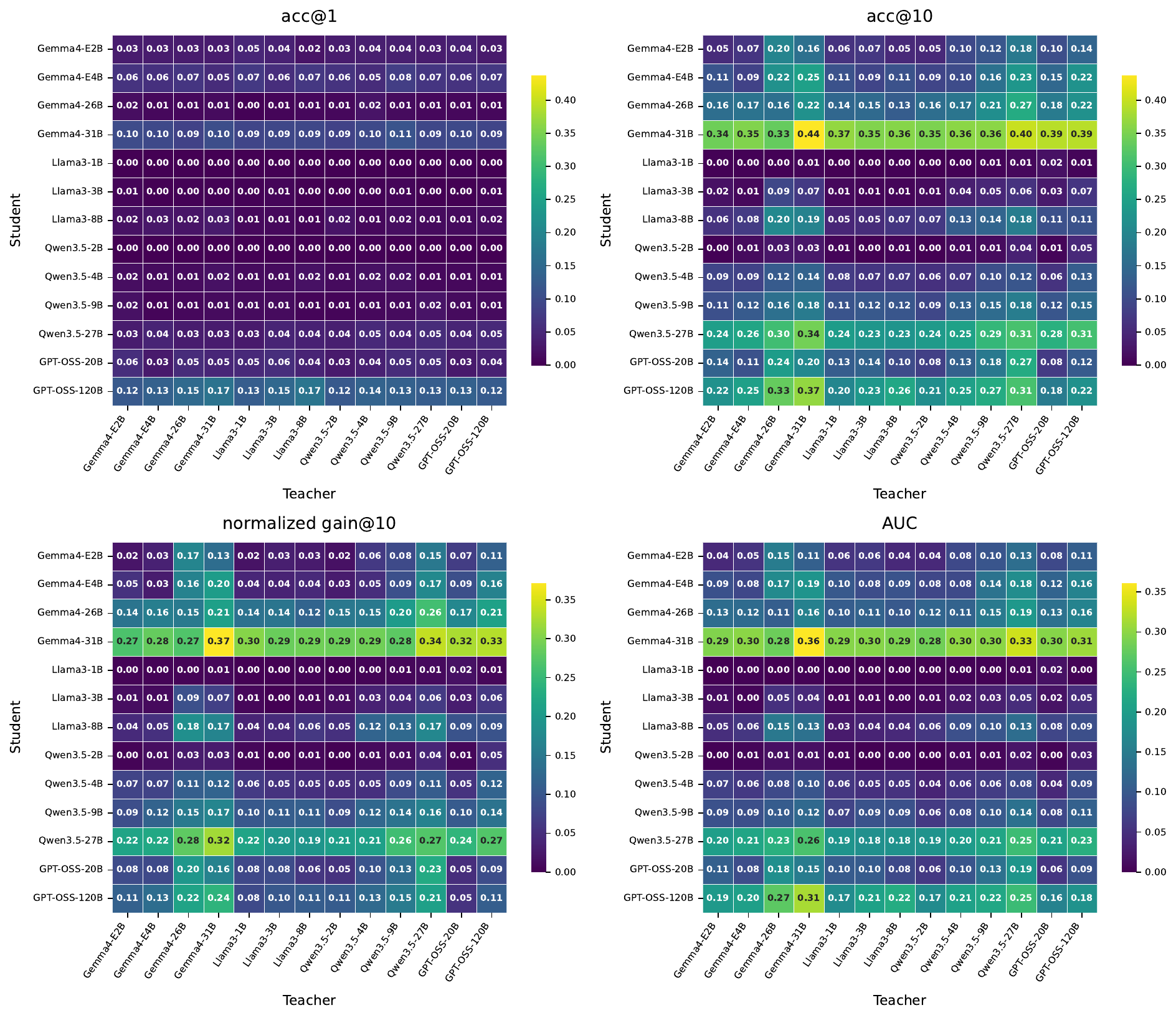}
\caption{ \textbf{BBEH Linguini dense model matrices}. Rows are students and columns are teachers.  We annotate all four metrics: $\acc@1$, $\acc@10$, $\gain$, and cumulative accuracy $\auc$.}
\label{fig:app-bbeh-linguini-matrix}
\end{figure}

\begin{figure}[ht]
\centering
\includegraphics[width=1.0\linewidth]{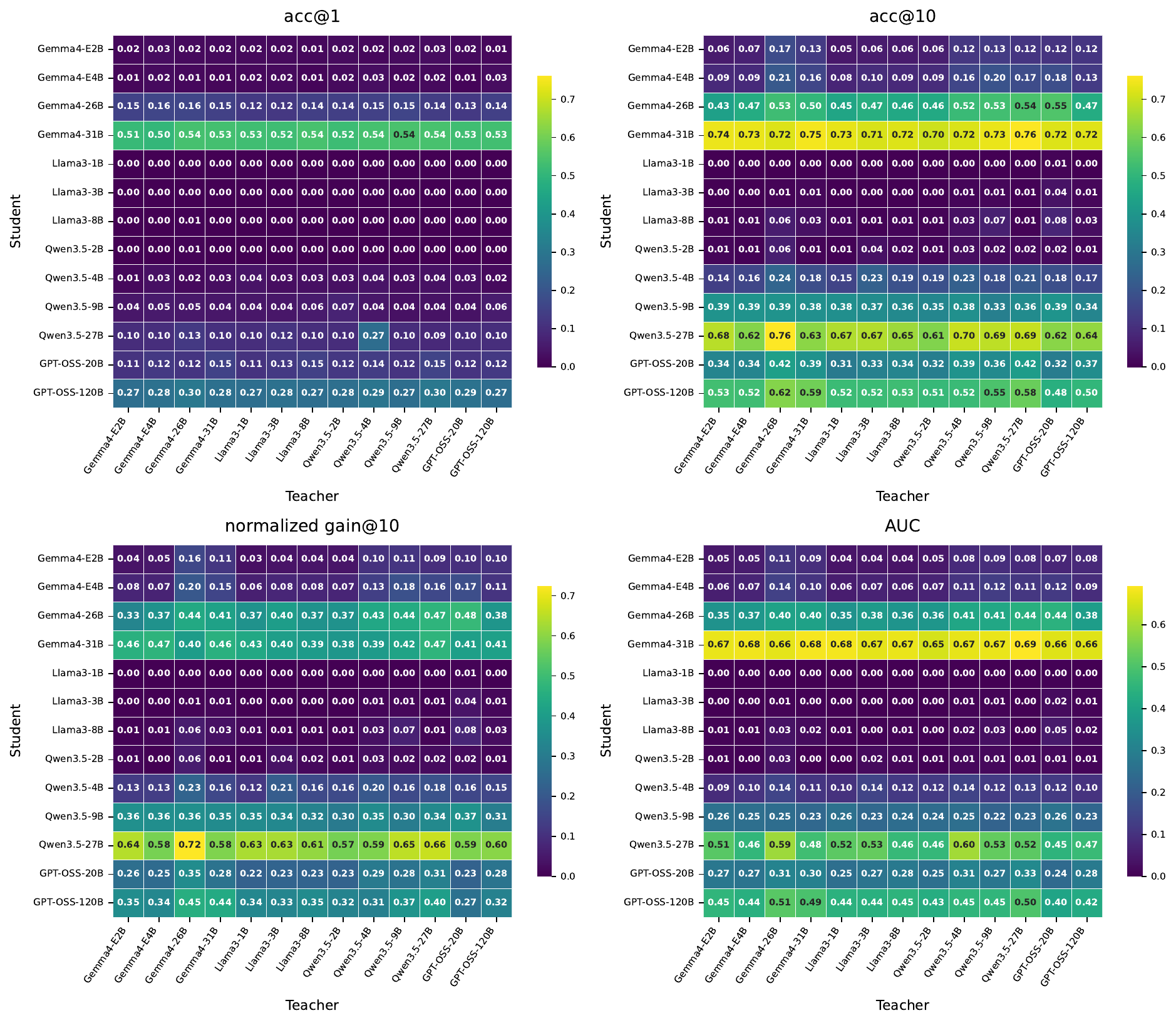}
\caption{\textbf{ARC-AGI1 dense model matrices}. Rows are students and columns are teachers.  We annotate all four metrics: $\acc@1$, $\acc@10$, $\gain$, and cumulative accuracy $\auc$.}
\label{fig:app-arc-agi1-matrix}
\end{figure}
\clearpage

\subsection{Supplementary Trajectory Figure}
\label{app:trajectory-figure}

\begin{figure}[ht]
\centering
\includegraphics[width=\linewidth]{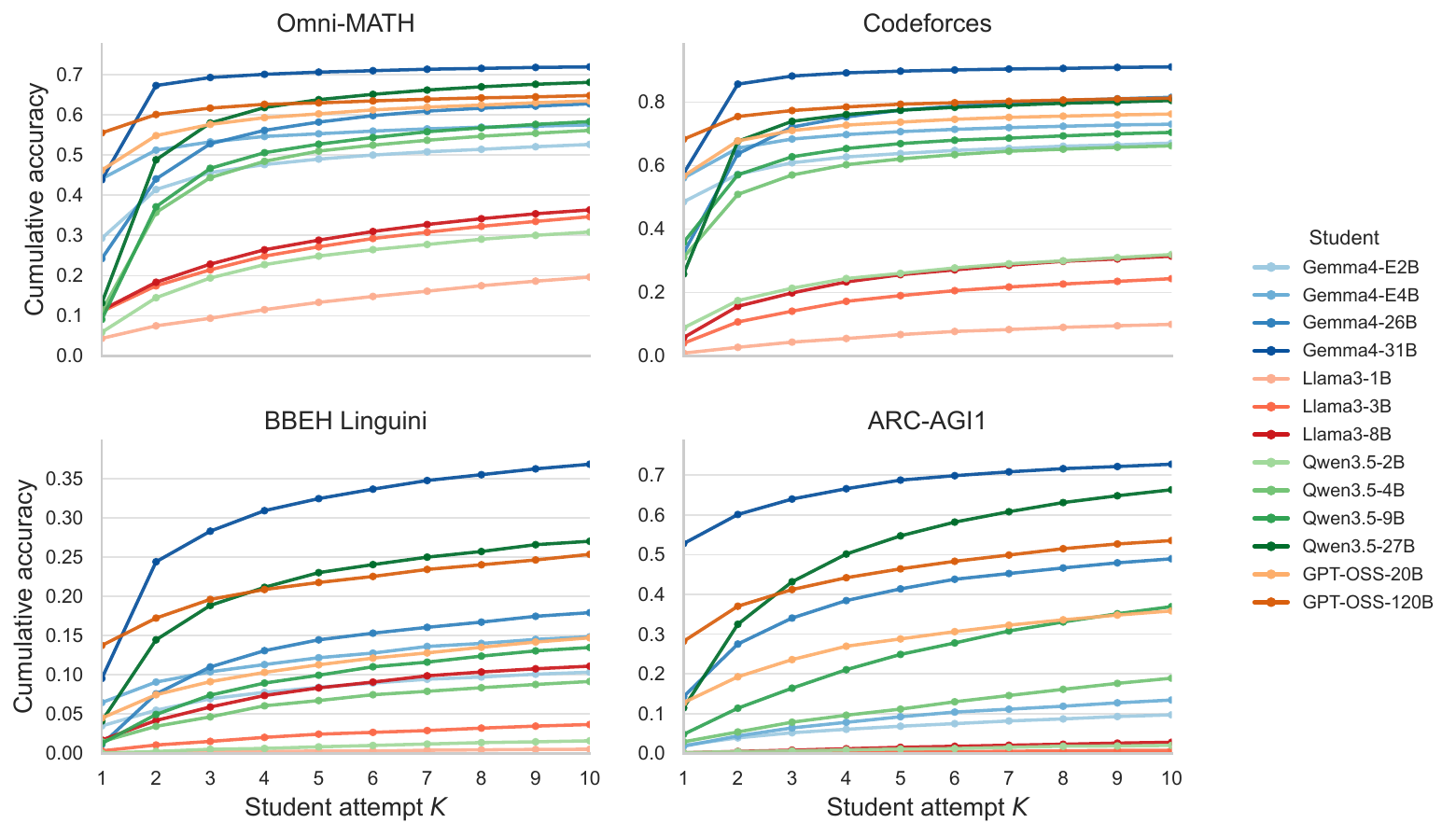}
\caption{\textbf{Cumulative accuracy curves are separated primarily by the
student model.}  Each panel shows one environment, and each curve averages
across teachers for one student.  The broad vertical separation between students
is larger than the within-student variation induced by changing teachers.}
\label{fig:student-cumulative-curves}
\end{figure}
\clearpage

\subsection{Supplementary Gain-by-Role Figures}
\label{app:gain-by-role-figures}

\begin{figure*}[ht]
\centering
\includegraphics[width=0.95\textwidth]{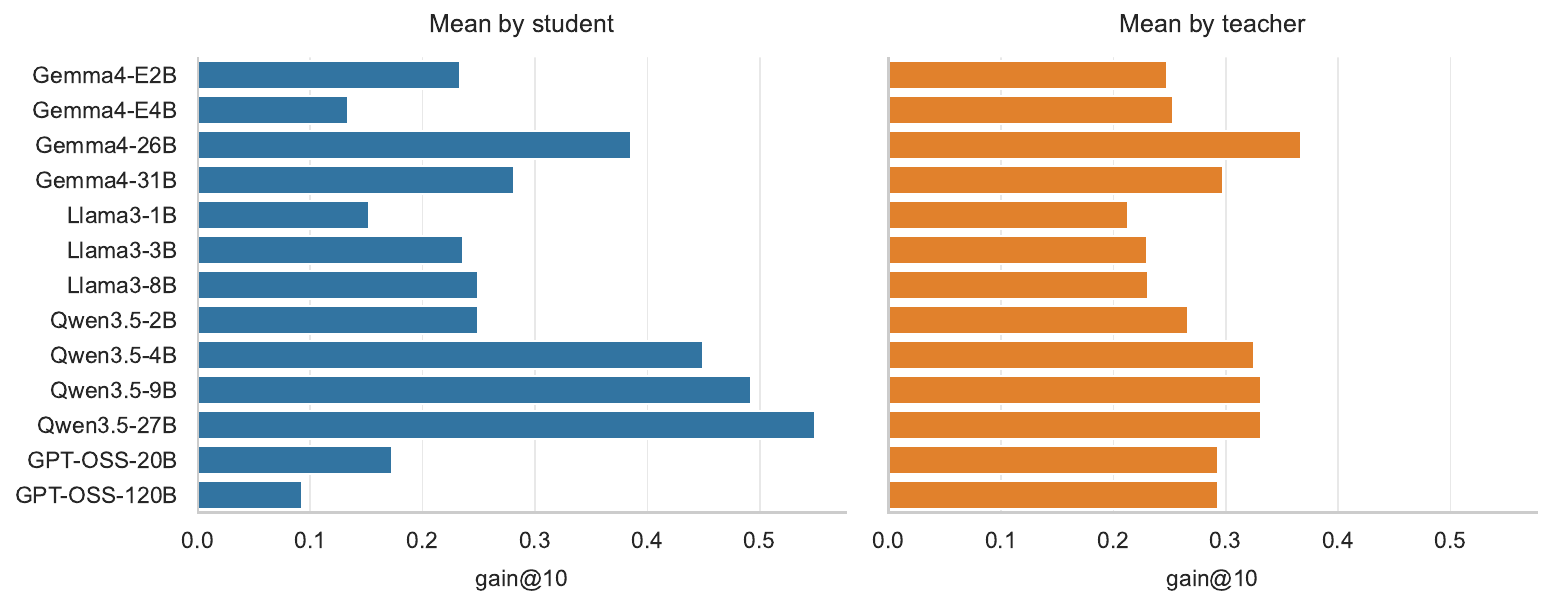}
\caption{Omni-MATH. We decompose gain by student(left) and teacher(right) for each model we compute  mean $\gain$ at $K=10$ across all teachers(left) and students(right).}
\label{fig:app-omni-math-gain-by-role}
\end{figure*}

\begin{figure*}[ht]
\centering
\includegraphics[width=0.95\textwidth]{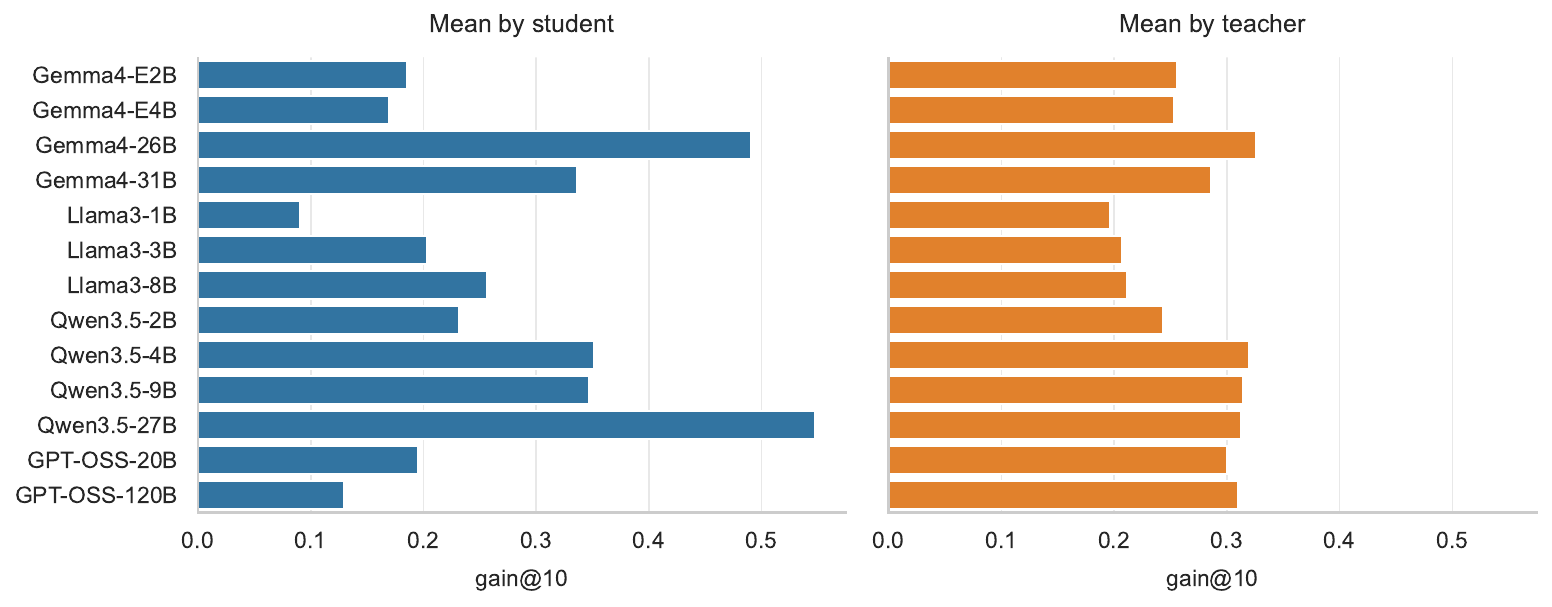}
\caption{Codeforces. We decompose gain by student(left) and teacher(right) for each model we compute  mean $\gain$ at $K=10$ across all teachers(left) and students(right).}
\label{fig:app-codeforces-gain-by-role}
\end{figure*}

\begin{figure*}[ht]
\centering
\includegraphics[width=0.95\textwidth]{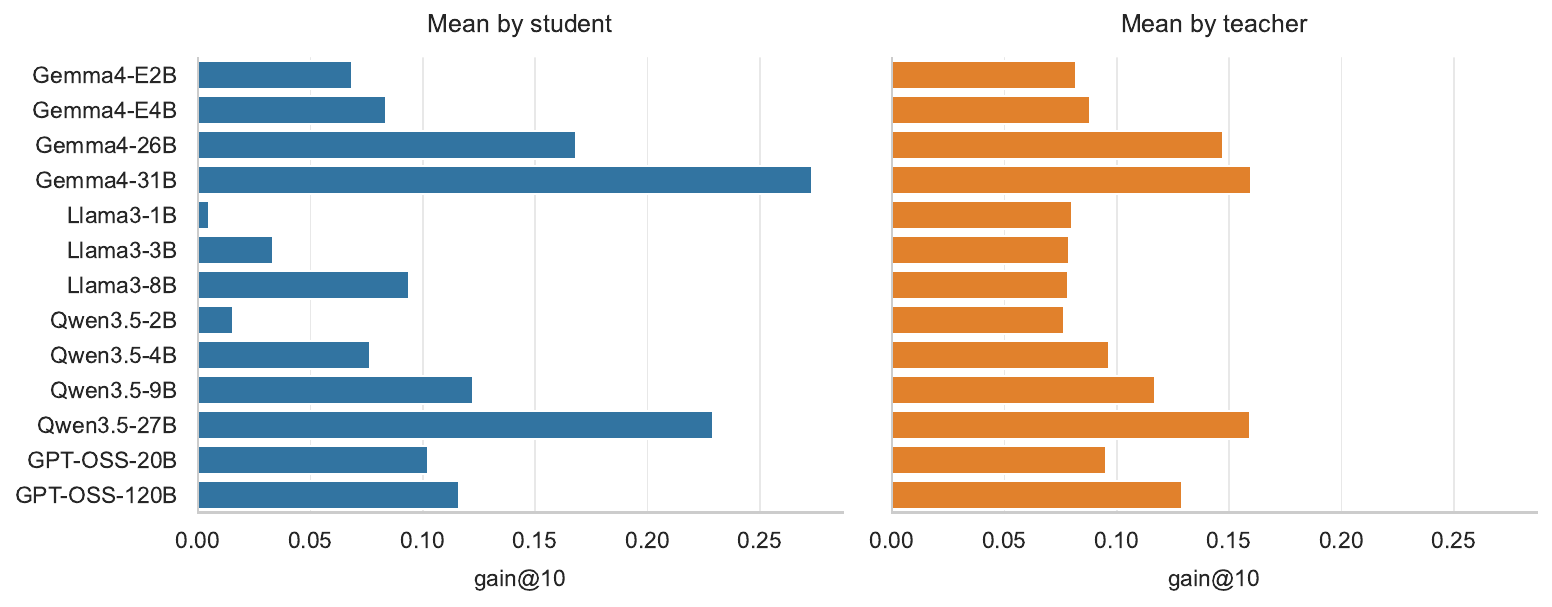}
\caption{BBEH Linguini. We decompose gain by student(left) and teacher(right) for each model we compute  mean $\gain$ at $K=10$ across all teachers(left) and students(right).}
\label{fig:app-bbeh-linguini-gain-by-role}
\end{figure*}

\begin{figure*}[ht]
\centering
\includegraphics[width=0.95\textwidth]{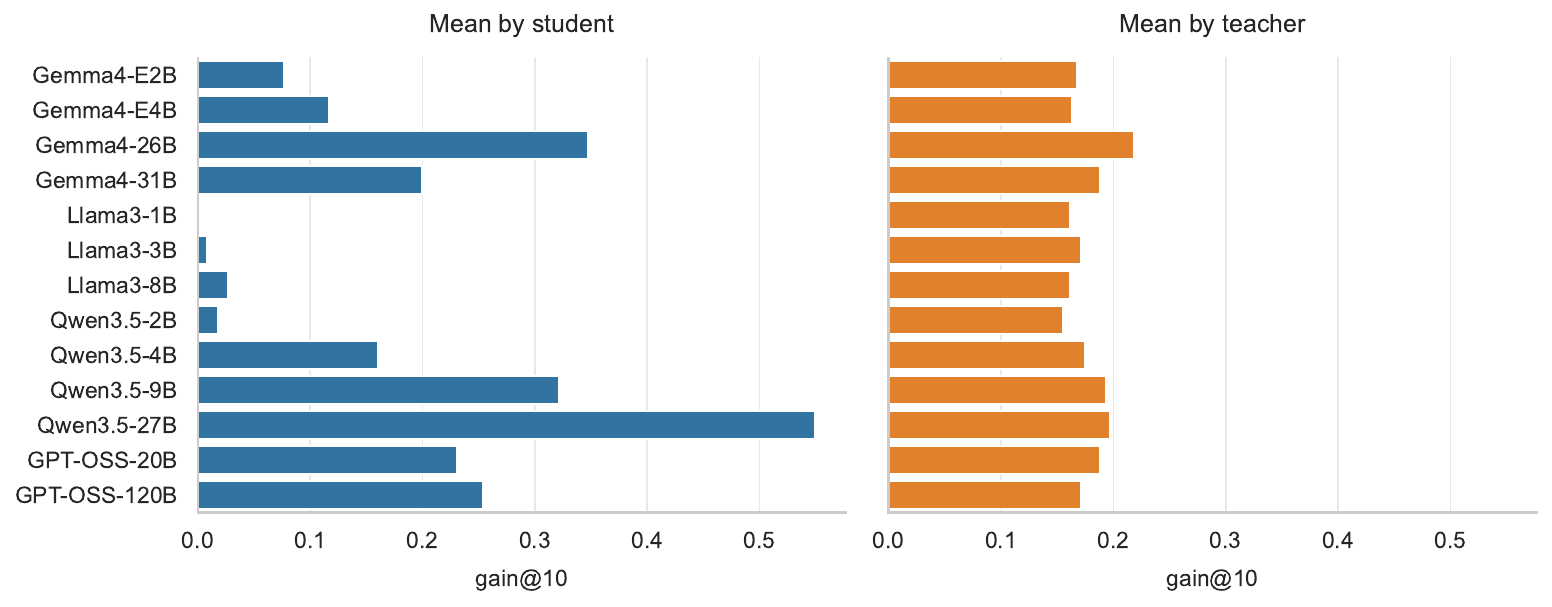}
\caption{ARC-AGI1. We decompose gain by student (left) and teacher (right) for each model we compute  mean $\gain$ at $K=10$ across all teachers(left) and students(right).}
\label{fig:app-arc-agi1-gain-by-role}
\end{figure*}
\clearpage
\subsection{Supplementary Gemma4 Ablation Figures}
\label{app:gemma4-ablation-figures}


\begin{figure}[ht]
\centering
\includegraphics[width=\linewidth]{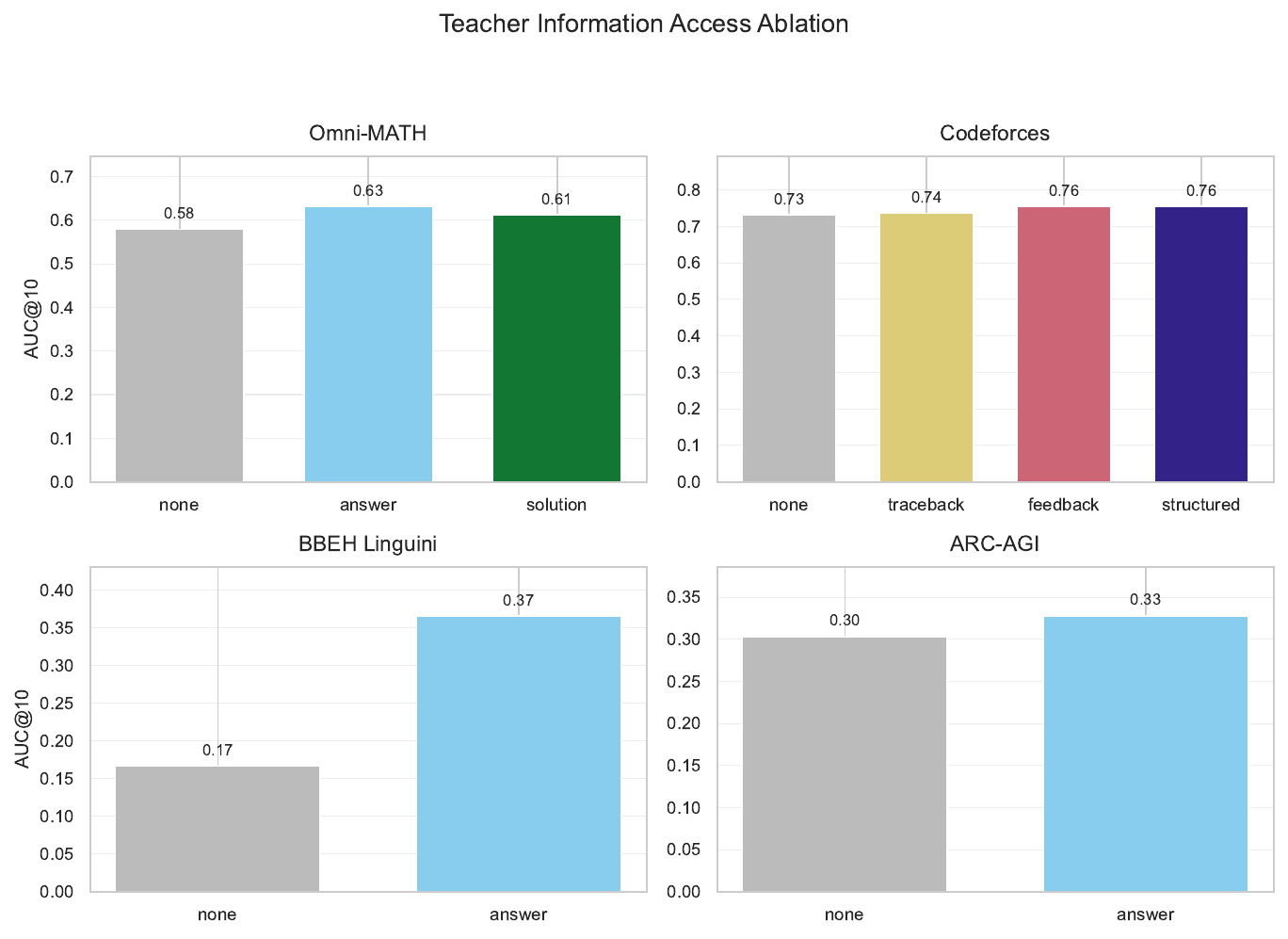}
\caption{\textbf{Gemma4 teacher-information access ablation across environments.}
Each bar reports cumulative accuracy $\auc$ at $K=10$ with max history 5,
averaged over Gemma4 student-teacher pair cells. Omni-MATH varies final-answer
and solution access, Codeforces varies execution-context access, and BBEH
Linguini and ARC-AGI1 compare no access against answer access.}
\label{fig:app-gemma4-teacher-access-ablation-full}
\end{figure}

\begin{figure}[ht]
\centering
\includegraphics[width=0.95\linewidth]{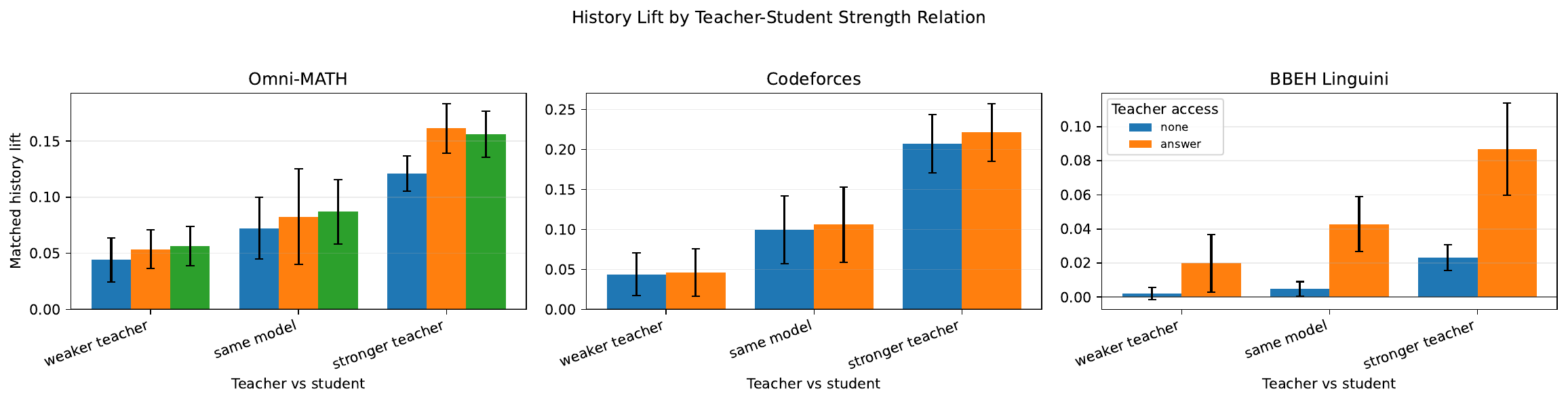}
\caption{\textbf{History lift by relative teacher capability.}
Matched $\auc$ lift from history 1 to history 5 is largest when the teacher is
stronger than the student.}
\label{fig:gemma4-history-lift-by-teacher-strength}
\end{figure}

\begin{figure}[ht]
\centering
\includegraphics[width=0.95\linewidth]{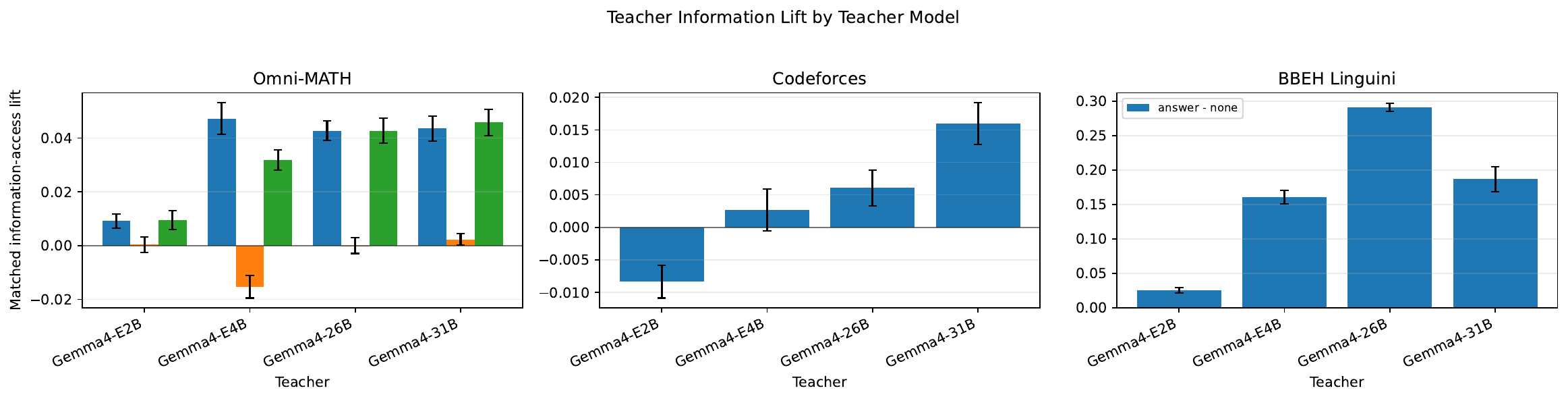}
\caption{\textbf{Information-access lift by teacher.}
Matched information-access gains vary by teacher and environment, rather than
appearing as a uniform benefit of revealing the answer or solution.}
\label{fig:gemma4-teacher-information-lift-by-teacher}
\end{figure}

\subsection{Feedback-Control Tables}
\label{app:feedback-control-tables}

\begin{table*}[p]
\centering
\caption{\textbf{Per-student controls disentangling interaction, feedback, and feedback quality.}
Cells report $\acc@10$ with $\gain@10$ in parentheses, in percentage points.
Self-refinement is the generic retry condition, self-feedback is the dense-matrix diagonal, and best feedback selects the highest-$\acc@10$ teacher available for the student and environment.}
\label{tab:feedback-control-full}
\scriptsize
\resizebox{\textwidth}{!}{%
\begin{tabular}{llcccc}
\toprule
Environment & Student & Self-refinement & Self-feedback & Best feedback & Best teacher \\
\midrule
Omni-MATH & Gemma4-E2B & 48.4 (+22.3) & 44.9 (+15.8) & 65.8 (+37.3) & Gemma4-26B \\
Omni-MATH & Gemma4-E4B & 56.6 (+11.5) & 52.9 (+7.0) & 66.4 (+22.1) & Gemma4-26B \\
Omni-MATH & Gemma4-26B & 49.8 (+26.4) & 67.2 (+43.2) & 68.0 (+43.8) & Qwen3.5-9B \\
Omni-MATH & Gemma4-31B & 66.0 (+23.6) & 71.5 (+27.0) & 72.9 (+31.1) & Qwen3.5-9B \\
Omni-MATH & Llama-1B & 11.1 (+7.4) & 8.8 (+4.7) & 28.3 (+24.0) & Gemma4-26B \\
Omni-MATH & Llama-3B & 22.9 (+11.3) & 22.9 (+13.3) & 46.7 (+35.4) & Gemma4-26B \\
Omni-MATH & Llama-8B & 24.6 (+15.2) & 20.7 (+10.0) & 51.8 (+40.8) & Gemma4-26B \\
Omni-MATH & Qwen3.5-2B & 22.3 (+16.4) & 25.6 (+19.9) & 41.6 (+35.0) & Gemma4-26B \\
Omni-MATH & Qwen3.5-4B & 40.0 (+28.9) & 59.6 (+48.8) & 65.4 (+54.3) & Gemma4-26B \\
Omni-MATH & Qwen3.5-9B & 34.4 (+24.6) & 62.9 (+53.1) & 68.0 (+58.4) & Gemma4-26B \\
Omni-MATH & Qwen3.5-27B & 61.7 (+49.6) & 72.5 (+59.0) & 72.5 (+59.0) & Qwen3.5-27B \\
Omni-MATH & GPT-OSS-20B & 65.1 (+17.6) & 60.5 (+13.3) & 70.1 (+23.4) & Gemma4-26B \\
Omni-MATH & GPT-OSS-120B & 67.8 (+14.3) & 61.5 (+6.6) & 69.7 (+15.0) & Gemma4-26B \\
Codeforces & Gemma4-E2B & 59.0 (+9.8) & 63.9 (+16.0) & 77.0 (+26.6) & Gemma4-26B \\
Codeforces & Gemma4-E4B & 66.8 (+12.5) & 69.7 (+13.9) & 80.3 (+25.4) & Gemma4-26B \\
Codeforces & Gemma4-26B & 75.8 (+44.1) & 86.7 (+52.5) & 86.7 (+52.5) & Gemma4-26B \\
Codeforces & Gemma4-31B & 84.2 (+27.1) & 91.4 (+32.4) & 93.0 (+35.2) & Qwen3.5-9B \\
Codeforces & Llama-1B & 1.8 (+1.0) & 1.6 (+0.8) & 21.3 (+20.1) & GPT-OSS-120B \\
Codeforces & Llama-3B & 13.7 (+8.5) & 8.4 (+4.1) & 36.5 (+33.6) & GPT-OSS-20B \\
Codeforces & Llama-8B & 14.5 (+8.0) & 12.5 (+7.8) & 45.9 (+40.6) & Qwen3.5-4B \\
Codeforces & Qwen3.5-2B & 19.5 (+10.5) & 26.4 (+18.2) & 41.6 (+33.4) & Qwen3.5-4B \\
Codeforces & Qwen3.5-4B & 65.3 (+28.0) & 68.6 (+39.1) & 76.4 (+43.4) & Gemma4-26B \\
Codeforces & Qwen3.5-9B & 64.6 (+22.1) & 72.3 (+36.3) & 77.0 (+40.8) & Gemma4-26B \\
Codeforces & Qwen3.5-27B & 71.9 (+39.0) & 81.4 (+55.3) & 86.7 (+60.4) & Gemma4-26B \\
Codeforces & GPT-OSS-20B & 72.3 (+11.3) & 75.2 (+19.5) & 84.4 (+27.0) & Gemma4-26B \\
Codeforces & GPT-OSS-120B & 77.5 (+9.4) & 82.4 (+14.1) & 86.5 (+17.8) & Gemma4-26B \\
BBEH Linguini & Gemma4-E2B & 11.0 (+7.0) & 5.3 (+2.0) & 19.7 (+16.3) & Gemma4-26B \\
BBEH Linguini & Gemma4-E4B & 16.7 (+9.0) & 8.7 (+2.7) & 24.7 (+19.3) & Gemma4-31B \\
BBEH Linguini & Gemma4-26B & 10.3 (+9.7) & 16.0 (+15.0) & 26.7 (+25.7) & Qwen3.5-27B \\
BBEH Linguini & Gemma4-31B & 32.1 (+19.3) & 43.7 (+33.3) & 43.7 (+33.3) & Gemma4-31B \\
BBEH Linguini & Llama-1B & 0.0 (+0.0) & 0.0 (+0.0) & 2.0 (+1.7) & GPT-OSS-20B \\
BBEH Linguini & Llama-3B & 1.7 (+1.3) & 0.7 (+0.0) & 9.3 (+9.0) & Gemma4-26B \\
BBEH Linguini & Llama-8B & 6.0 (+4.0) & 6.7 (+6.0) & 20.3 (+18.0) & Gemma4-26B \\
BBEH Linguini & Qwen3.5-2B & 0.3 (+0.3) & 0.0 (+0.0) & 5.0 (+5.0) & GPT-OSS-120B \\
BBEH Linguini & Qwen3.5-4B & 5.3 (+3.7) & 7.0 (+5.0) & 13.7 (+11.7) & Gemma4-31B \\
BBEH Linguini & Qwen3.5-9B & 7.3 (+5.7) & 15.3 (+14.0) & 18.3 (+16.0) & Qwen3.5-27B \\
BBEH Linguini & Qwen3.5-27B & 22.7 (+19.0) & 31.0 (+25.7) & 34.3 (+31.0) & Gemma4-31B \\
BBEH Linguini & GPT-OSS-20B & 15.0 (+8.0) & 7.7 (+4.7) & 26.7 (+21.3) & Qwen3.5-27B \\
BBEH Linguini & GPT-OSS-120B & 27.7 (+12.3) & 22.0 (+9.7) & 36.7 (+20.0) & Gemma4-31B \\
ARC-AGI1 & Gemma4-E2B & 5.8 (+3.8) & 5.5 (+3.8) & 17.2 (+15.2) & Gemma4-26B \\
ARC-AGI1 & Gemma4-E4B & 8.8 (+7.0) & 9.0 (+6.8) & 21.0 (+19.5) & Gemma4-26B \\
ARC-AGI1 & Gemma4-26B & 38.0 (+24.8) & 52.5 (+36.8) & 55.0 (+42.3) & GPT-OSS-20B \\
ARC-AGI1 & Gemma4-31B & 21.5 (+11.0) & 74.8 (+21.5) & 75.5 (+21.8) & Qwen3.5-27B \\
ARC-AGI1 & Llama-1B & 0.0 (+0.0) & 0.2 (+0.2) & 0.8 (+0.8) & GPT-OSS-20B \\
ARC-AGI1 & Llama-3B & 0.0 (+0.0) & 0.0 (+0.0) & 3.8 (+3.8) & GPT-OSS-20B \\
ARC-AGI1 & Llama-8B & 1.0 (+0.8) & 1.0 (+0.8) & 7.8 (+7.8) & GPT-OSS-20B \\
ARC-AGI1 & Qwen3.5-2B & 0.5 (+0.5) & 1.0 (+1.0) & 6.2 (+5.5) & Gemma4-26B \\
ARC-AGI1 & Qwen3.5-4B & 10.5 (+8.5) & 23.0 (+19.0) & 24.5 (+22.5) & Gemma4-26B \\
ARC-AGI1 & Qwen3.5-9B & 28.0 (+24.3) & 32.5 (+29.0) & 39.2 (+35.5) & GPT-OSS-20B \\
ARC-AGI1 & Qwen3.5-27B & 36.8 (+27.2) & 69.2 (+60.5) & 76.0 (+63.3) & Gemma4-26B \\
ARC-AGI1 & GPT-OSS-20B & 33.8 (+22.3) & 31.5 (+20.0) & 42.5 (+30.8) & Gemma4-26B \\
ARC-AGI1 & GPT-OSS-120B & 52.2 (+20.5) & 50.0 (+23.0) & 62.0 (+31.5) & Gemma4-26B \\
\bottomrule
\end{tabular}%
}
\end{table*}


\end{document}